\documentclass[letterpaper]{article} 
\usepackage{aaai23}  
\usepackage{times}  
\usepackage{helvet}  
\usepackage{courier}  
\usepackage[hyphens]{url}  
\usepackage{graphicx} 
\usepackage{natbib}  
\usepackage{caption} 
\usepackage{algorithm}
\usepackage{algorithmic}

\usepackage{rotating}
\usepackage{adjustbox}

\usepackage{newfloat}
\usepackage{listings}
\DeclareCaptionStyle{ruled}{labelfont=normalfont,labelsep=colon,strut=off} 
\floatstyle{ruled}
\newfloat{listing}{tb}{lst}{}
\floatname{listing}{Listing}
\usepackage{amsmath,amsfonts,bm}

\DeclareMathOperator{\diag}{diag}
\newcommand{\esv}[1]{^{(#1)}} 

\def\eqref#1{equation~\ref{#1}}

\def\1{\bm{1}}

\def\vtheta{{\bm{\theta}}}

\def\vc{{\bm{c}}}

\def\ve{{\bm{e}}}

\def\vh{{\bm{h}}}

\def\vs{{\bm{s}}}

\def\vw{{\bm{w}}}
\def\vx{{\bm{x}}}
\def\vy{{\bm{y}}}

\def\mH{{\bm{H}}}

\def\mW{{\bm{W}}}
\def\mX{{\bm{X}}}

\DeclareMathAlphabet{\mathsfit}{\encodingdefault}{\sfdefault}{m}{sl}
\SetMathAlphabet{\mathsfit}{bold}{\encodingdefault}{\sfdefault}{bx}{n}

\def\gY{{\mathcal{Y}}}

\def\sR{{\mathbb{R}}}

\usepackage{todonotes}
\usepackage{booktabs}
\usepackage{amsfonts}
\usepackage{xcolor}
\usepackage{breqn}

\title{Weight Predictor Network with Feature Selection \\ for Small Sample Tabular Biomedical Data}
\author{
    Andrei Margeloiu, Nikola Simidjievski, Pietro Li\'{o}, Mateja Jamnik
}
\affiliations{
    
    University of Cambridge\\
    
    \{am2770, ns779, pl219, mj201\}@cam.ac.uk
}

\begin{document}

\maketitle

\begin{abstract}
Tabular biomedical data is often high-dimensional but with a very small number of samples. Although recent work showed that well-regularised simple neural networks could outperform more sophisticated architectures on tabular data, they are still prone to overfitting on tiny datasets with many potentially irrelevant features. To combat these issues, we propose \textbf{W}eight \textbf{P}redictor Network with \textbf{F}eature \textbf{S}election (\textbf{WPFS}) for learning neural networks from high-dimensional and small sample data by reducing the number of learnable parameters and simultaneously performing feature selection. In addition to the classification network, WPFS uses two small auxiliary networks that together output the weights of the first layer of the classification model. We evaluate on nine real-world biomedical datasets and demonstrate that WPFS outperforms other standard as well as more recent methods typically applied to tabular data. Furthermore, we investigate the proposed feature selection mechanism and show that it improves performance while providing useful insights into the learning task.
\end{abstract}

\section{Introduction}

The wide availability of high-throughput sequencing technologies has enabled the collection of high-dimensional biomedical data containing measurements of the entire genome. However, many clinical trials which collect such high-dimensional, typically tabular, data are carried out on small cohorts of patients due to the scarcity of available patients, prohibitive costs, or the risks associated with new medicines. Although using deep networks to analyse such feature-rich data promises to deliver personalised medicine, current methods tend to perform poorly when the number of features significantly exceeds the number of samples.

Recently, a large number of neural architectures have been developed for tabular data, taking inspiration from tree-based methods \citep{katzir2020net, hazimeh2020tree, Popov2020NeuralOD, Yang2018DeepND}, including attention-based modules \citep{arik2021tabnet, Huang2020TabTransformerTD}, explicitly performing feature selection \citep{Yang2021LocallySN, lemhadri2021lassonet, yoon2018invase}, or modelling multiplicative feature interactions \citep{Qin2021AreNR}. However, such specialised neural architectures are not necessarily superior to tree-based ensembles (such as gradient boosting decision trees) on tabular data, and their relative performance can vary greatly between datasets \citep{Gorishniy2021RevisitingDL}.

On tabular data, specialised neural architectures are outperformed by well-regularised feed-forward networks such as MLPs \citep{kadra2021well}. As such, we believe that designing regularisation mechanisms for MLPs holds great potential for improving performance on high-dimensional, small size tabular datasets.

\begin{table*}[t]
\centering
\small
\begin{tabular}{lccccccc}
\toprule
\textbf{Method}              & \textbf{Data Type}$^1$   & \textbf{End-to-end} & \textbf{Feature} & \textbf{Additional}  & \textbf{Additional Training}               & \textbf{Feature}   &  \textbf{Differentiable}     \\
&    & \textbf{Training}   & \textbf{Selection}                 & \textbf{Decoder}     & \textbf{Hyper-parameters}            & \textbf{Embedding} & \textbf{Loss} \\ \midrule

DietNetworks        & Discrete   & \checkmark        & $\times$                  & \checkmark         & 1 & Histogram   &   \checkmark  \\                
FsNet      & Cont. & \checkmark & \checkmark & \checkmark      & 2   & Histogram & \checkmark          \\ 
CAE        & Cont. & \checkmark & \checkmark & $\times$ &  1       & -  &    \checkmark              \\

LassoNet           & Cont. & \checkmark          & \checkmark &   $\times$         &      2       & -    &  $\times$       \\
DNP & Cont. & $\times$   & \checkmark &    $\times$        &           1                  & -    & \checkmark \\
SPINN & Cont. &    \checkmark       & \checkmark  & $\times$        &      4       &  -      & $\times$       \\ \midrule

\textbf{WPFS (ours)} & \textbf{Cont.} & \checkmark & \checkmark & $\times$       & \textbf{1}         & \textbf{Matrix factor.} & \checkmark \\
\bottomrule
\end{tabular}

\caption{\textit{Comparison with Related Work}. $^1$Denotes the most general regime under which each method is appropriate, but note that all methods for continuous data also support discrete data.}
\label{tab:related-work}

\end{table*}

An important characteristic of feed-forward neural networks applied to high-dimensional data is that the first hidden layer contains a disproportionate number of parameters compared to the other layers. For example, on a typical genomics dataset with $\sim {20,\!000}$ features, a five-layer MLP with $100$ neurons per layer contains $98\%$ of the parameters in the first layer. We hypothesise that having a large learning capacity in the first layer contributes overfitting of neural networks on high-dimensional, small sample datasets.

In this paper, we propose \textbf{W}eight \textbf{P}redictor Network with \textbf{F}eature \textbf{S}election (\textbf{WPFS}), a simple but effective method to overcome overfitting on high-dimensional, small sample tabular supervised tasks. Our method uses two tiny auxiliary networks to perform global feature selection and compute the weights of the first hidden layer of a classification network. Our method is generalisable and can be applied to any neural network in which the first hidden layer is linear. More specifically, our contributions\footnote{Code is available at \url{https://github.com/andreimargeloiu/WPFS}} include:


\begin{enumerate}
    \item A novel method to substantially reduce the number of parameters in feed-forward neural networks and simultaneously perform global feature selection (Sec.~\ref{sec:method}). Our method extends DietNetworks to support continuous data, use factorised feature embeddings, perform feature selection, and it consistently improves performance.
    \item A novel global feature selection mechanism that ranks features during training based on feature embeddings (Sec. \ref{sec:method}). We propose a simple and differentiable auxiliary loss to select a small subset of features (Sec. \ref{sec:training-objective}). We investigate the proposed mechanisms and show that they improve performance (Sec. \ref{sec:influence-spn-performance}) and provide insights into the difficulty of the learning task (Sec. \ref{sec:spn-interpretability}).
    \item We evaluate WPFS on $9$ high-dimensional, small sample biomedical datasets and demonstrate that it performs better overall than $10$ methods including TabNet, DietNetworks, FsNet, CAE, MLPs, Random Forest, and Gradient Boosted Trees (Sec. \ref{sec:predictive-performance}). We attribute the success of WPFS to its ability to reduce overfitting (Sec.~\ref{sec:training-behaviour}).
\end{enumerate}

\section{Related Work}
\label{sec:rw}

We focus on learning problems from tabular datasets where the number of features substantially exceeds the number of samples. As such, this work relates to DietNetworks \citep{Romero2017DietNT}, which uses auxiliary networks to predict the weights of the first layer of an underlying feed-forward network (referred to as ``fat layer''). FsNet \citep{singh2020fsnet} extends DietNetworks by combining them with Concrete autoencoders (CAE) \citep{balin2019concrete}. DietNetworks and FsNet use histogram feature embeddings, which are well-suited for discrete data, and require a decoder (with an additional reconstruction loss) to alleviate training instabilities. These mechanisms do not translate well to continuous-data problems (Sec. ~\ref{sec:feature-embedding-performance}). In contrast, our WPFS takes matrix-factorised feature embeddings and uses a Sparsity Network for global feature selection, which help alleviate training instabilities, remove the need for an additional decoder network, and improve performance (Sec.~\ref{sec:training-behaviour}). Table \ref{tab:related-work} further highlights the key differences between WPFS and the discussed related methods.

When learning from small sample tabular datasets, various neural architectures use a regularisation mechanism, such as $L_1$ \citep{tibshirani1996lasso} for performing feature selection. However, optimising $L_{1}$-constrained problems in parameter-rich neural networks may lead to local sub-optima \citep{bach2017breaking}. To combat this, \citet{liu2017deep} propose DNPs which employ a greedy approximation of $L_1$ focusing on individual features. In a similar context, SPINN \citep{Feng2017SparseInputNN} employ a sparse group lasso regularisation \citep{simon2013sparse} on the first layer. LassoNet \citep{lemhadri2021lassonet} combines a linear model (with $L_1$ regularisation) with a non-linear neural network to perform feature selection. These methods, however, typically require lengthy training procedures; be it retraining the model multiple times (DNP), using specialised optimisers (LassoNet, SPINN), or costly hyper-parameter optimisation (LassoNet, SPINN). All of these may lead to unfavourable properties when learning from high-dimensional, small size datasets.

In a broader context, approaches to reducing the number of learnable parameters have primarily addressed image-data problems rather than tabular data. Hypernetworks \citep{Ha2017HyperNetworks} use an external network to generate the entire weight matrix for each layer of a classification network. However, the first layer is very large on high-dimensional tabular data and requires using large Hypernetworks. Also, Hypernetworks learn an embedding per layer, which can overfit on small size datasets. In contrast, our WPFS computes the weight matrix column-by-column from unsupervised feature embeddings.

\begin{figure*}[t!]
    \centering
    \includegraphics[width=0.87\textwidth]{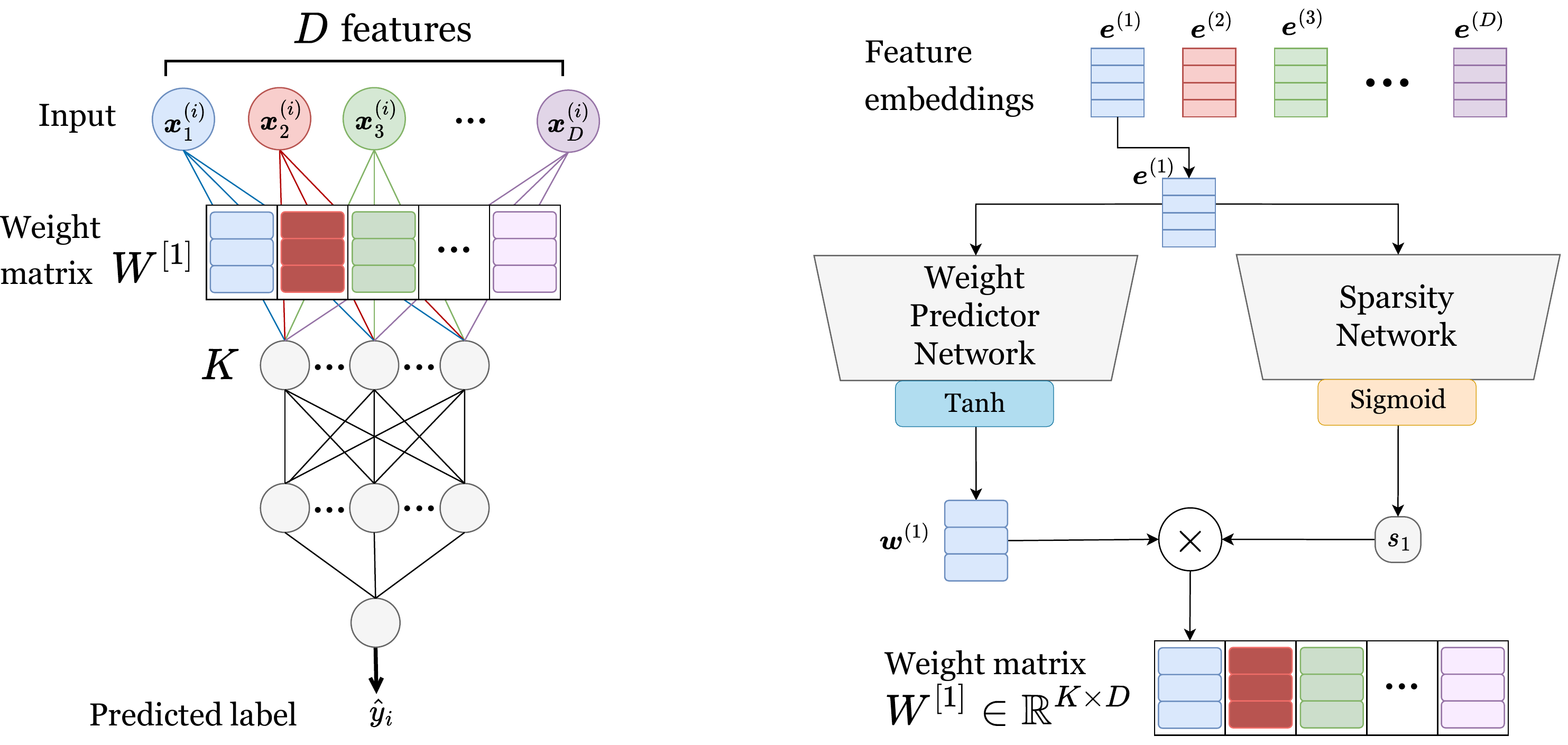}
    \caption{Architecture of the proposed \textbf{W}eight \textbf{P}redictor Network with \textbf{F}eature \textbf{S}election (\textbf{WPFS}). (Left) A standard feed-forward neural network, highlighting the weight matrix $\mW^{[1]} \in \sR^{K \times D}$ of the first layer. (Right) We compute feature embeddings $\ve^{(j)}$ for each feature $j$. WPFS uses two auxiliary networks to compute $\mW^{[1]}$. First, the Weight Predictor Network (WPN) maps each feature embedding $\ve^{(j)}$ to a vector $\vw^{(j)}$. The Sparsity Network (SPN) maps the same feature embeddings $\ve^{(j)}$ to scalars $s_j \in (0, 1)$. The two outputs are multiplied $\vw^{(j)} \cdot s_j$, and the result acts as the $j$-th column of $\mW^{[1]}$. An element $\mW^{[1]}_{k, j}$ represents the weight between feature $j$ and the neuron $k$ from the first layer. All three networks are trained simultaneously.}
    \label{fig:architecture}
\end{figure*}

\section{Method}
\label{sec:method}
\subsection{Problem Formulation and Notation}

We consider the problem setting of classification in $\gY$ classes. Let $\{ (\vx^{(i)}, y_i) \}_{i=1}^N$ be a $D$-dimensional dataset of $N$ samples, where $\vx^{(i)} \in \sR^D$ (continuous) is a sample and $y_i \in \gY$ (categorical) is its label. The $j$-th feature of the $i$-th sample is represented as $\vx^{(i)}_j$. We define the data matrix as $\mX := [\vx^{(1)}, ..., \vx^{(N)}]^\top \in \sR^{N \times D}$, and the labels as $\vy:= [y_1, ..., y_N]$. For each feature $j$ we consider having an embedding $\ve^{(j)} \in \sR^M$ of size $M$ (Sections \ref{sec:feature-embedding-description}, \ref{sec:feature-embedding-performance} explain and analyse multiple embedding types).

Let $f_{\mW}: \sR^D \rightarrow \gY$ denote a classification neural network with parameters $\mW$, which inputs a sample $\vx^{(i)}$ and outputs a predicted label $\hat{y}_i$. In this paper we consider only the class of neural network in which the first layer is linear. Let the first layer have size $K$ and define $\mW^{[1]} \in \sR^{K \times D}$ to be its weight matrix.

\subsection{Model}
We propose \textbf{WPFS} as an approach for generating the weights of the first layer in a neural network. Instead of learning $\mW^{[1]}$ directly, WPFS uses two small auxiliary networks to compute $\mW^{[1]}$ column-by-column (Figure~\ref{fig:architecture}). We want to bypass learning $\mW^{[1]}$ directly because on high-dimensional tasks it contains over $95\%$ of the model's learnable parameters, which increases the risk of overfitting. In practice, the proposed method reduces the total number of learnable parameters by $\sim 90\%$.

Next, we describe the two auxiliary networks. The pseudocode is summarised in Algorithm \ref{alg:algorithm-pseudocode}.

The \textbf{W}eight \textbf{P}redictor \textbf{N}etwork (\textbf{WPN}) $f_{\vtheta_{\text{WPN}}}: \sR^{M} \rightarrow \sR^K$, with learnable parameters $\vtheta_{\text{WPN}}$, maps a feature embedding $\ve \esv{j}$ to a vector $\vw \esv{j} \in \sR^K$. Intuitively, for every feature $j$, the WPN outputs the weights $\vw^{(j)}$ connecting feature $j$ with all $K$ neurons in the first hidden layer. All outputs are concatenated horizontally in the matrix $\mW_{\text{WPN}} := [\vw \esv{1}, ..., \vw \esv{D}] \in \sR^{K \times D}$. We implement WPN as a $4$-layer MLP with a $tanh$ activation in the last layer.

The \textbf{Sp}arsity \textbf{N}etwork (\textbf{SPN}) $f_{\vtheta_{\text{SPN}}}: \sR^{M} \rightarrow \sR$, with learnable parameters $\vtheta_{\text{SPN}}$, approximates the global feature importance scores. For every feature $j$, the SPN maps a feature embedding $\ve \esv{j}$ to a scalar $s_j \in (0, 1)$ which represents the global feature importance score of feature $j$. A feature has maximal importance if $s_j \rightarrow 1$, and can be dropped if $s_j \rightarrow 0$. All outputs are stored into a vector $\vs := [s_1, ..., s_D] \in \sR^{D}$. We implement SPN as a $4$-layer MLP with a $sigmoid$ activation in the last layer. 

The outputs of WPN and SPN are combined to compute the weight matrix of first linear layer of the classification model as $\mW^{[1]} := [\vw \esv{1} \cdot s_1, ..., \vw \esv{D} \cdot s_D] = \mW^{[1]}_{\text{WPN}} \diag(\vs)$ (where $\diag(\vs)$ is a square diagonal matrix with the vector $\vs$ on the main diagonal). The last equation provides two justifications for the SPN.

On the one hand, the SPN can be interpreted as performing global feature selection. Passing an input $\vx$ through the first layer is equivalent to $\mW^{[1]} \vx = \mW^{[1]}_{\text{WPN}} \diag(\vs) \vx = \mW^{[1]}_{\text{WPN}} (\diag(\vs) \vx) = \mW^{[1]}_{\text{WPN}} (\vs \odot \vx)$ (where $\odot$ is the element-wise multiplication, also called the Hadamard product). The last equation illustrates that the feature importance scores $\vs$ act as a `mask' applied to the input $\vx$. A feature $j$ is ignored when $s_j \rightarrow 0$. In practice, SPN learns to ignore most features (see Sec. \ref{sec:spn-interpretability}).

On the other hand, the SPN can be viewed as rescaling $\mW^{[1]}$ because each feature importance score $s_j$ scales column $j$ of the matrix $\mW^{[1]}$. This interpretation solves a major limitation of DietNetwork, which uses a $tanh$ activation to compute the weights, and in practice, $tanh$ usually saturates and outputs large values $\in \{ -1, 1 \}$. Note that SPN enables learning weights with small magnitudes, as is commonly used in successful weight initialisation schemes \citep{glorot2010understanding, he2015delving}.

\subsection{Feature Embeddings}
\label{sec:feature-embedding-description}
We investigate four methods to obtain feature embeddings $\ve^{( )}$, which serve as input to the auxiliary networks (WPN and SPN). We assume no access to external data or feature embeddings because we want to simulate real-world medical scenarios in which feature embeddings are usually unavailable. Previous work \citep{Romero2017DietNT} showed that learning feature embeddings using end-to-end training, denoising autoencoders, or random projection may not be effective on small datasets, and we do not consider such embeddings. 

We consider only unsupervised methods for computing feature embeddings because of the high risk of overfitting on supervised tasks with small training datasets:

\begin{enumerate}
    \item \textbf{Dot-histogram} \citep{singh2020fsnet} embeds the shape and range of the distribution of the measurements for a feature. For every feature $j$, it computes the normalised histogram of the feature value $\mX_{:, j}$. If we consider $\vh^{(j)}, \vc^{(j)}$ to represent the heights and the centers of the histogram's bins, the dot-histogram embedding is the element-wise product $\ve^{(j)} := \vh^{(j)} \odot \vc^{(j)}$.
    \item \textbf{`Feature-values'} defines the embedding $\ve^{(j)} := \mX_{:, j}$ (i.e., $\ve^{(j)}$ contains the measurements for feature $j$). This relatively simple embedding uses the scarcity of training samples to its advantage by preserving all information.
    \item \textbf{Singular value decomposition} (\textbf{SVD}) has been used to compare genes expressions across different arrays \cite{alter2000singular}. It factorises the data matrix $\mX = \text{eigengenes} \times \text{eigenarrays}$, where the ``eigengenes'' define an orthogonal space of representative gene prototypes and the ``eigenarrays" represent the coordinates of each gene in this space. An embedding $\ve^{(j)}$ is the $j$-th column of the ``eigenarrays" matrix.
    \item \textbf{Non-negative matrix factorisation} (\textbf{NMF}) has been applied in bioinformatics to cluster gene expression \citep{kim2007sparse, taslaman2012framework} and identify common cancer mutations \citep{alexandrov2013deciphering}. It approximates $\mX \approx \mW \mH$, with the intuition that the column space of $\mW$ represents ``eigengenes'', and the column $\mH_{:, j}$ represents coordinates of gene $j$ in the space spanned by the eigengenes. The feature embedding is $\ve^{(j)} := \mH_{:, j}$.
\end{enumerate}

We found that the data matrix $\mX$ must be preprocessed using min-max scaling in $[0, 1]$ before computing the embeddings (Appx. \ref{appendix:embedding-preprocessing}). We analyse the performance of each embedding type in Section \ref{sec:feature-embedding-performance}.

\begin{algorithm}[t!]
\caption{Training the proposed method WPFS}
\label{alg:algorithm-pseudocode}
\small
\textbf{Input}: training data $\mX \in \sR^{N \times D}$, training labels $\vy \in \sR^N$, classification network $f_{\mW}$, weight predictor network $f_{\theta_{\text{WPN}}}$, sparsity network $f_{\theta_{\text{SPN}}}$, sparsity loss hyper-parameter $\lambda$, learning rate $\alpha$

\begin{algorithmic}[t!] 

\STATE Compute global feature embeddings:\\\hspace{\algorithmicindent} $\ve^{(1)}, \ve^{(2)}, ..., \ve^{(D)}$ from $\mX$

\FOR{each minibatch $ B =\{ (\vx^{(i)}, y_i) \}_{i=1}^b$ }
    \FOR{each feature $j = 1, 2, ..., D$ } 
    \STATE $\vw^{(j)} = f_{\theta_{\text{WPN}}}(\ve^{(j)})$ \COMMENT{{\scriptsize Pass feature embedding through the WPN}} \\
    \STATE $s_j = f_{\theta_{\text{SPN}}}(\ve^{(j)})$ \COMMENT{{\scriptsize Pass feature embedding through the SPN}} \\
    \STATE $\vw^{(j)} = \vw^{(j)} \cdot s_j$\\
    \ENDFOR
    \STATE Let $\mW^{[1]}$ to be the matrix $\in \sR^{K \times D}$ obtained by\\\hspace{\algorithmicindent} horizontally concatenating $\vw^{(1)}, \vw^{(2)}, ..., \vw^{(D)}$ \\
    
    \STATE Make $\mW^{[1]}$ the weight matrix of the first layer of $f_{\mW}$

    \FOR{each sample $i = 1, 2, ..., b$}
    \STATE $\hat{y}_i = f_{\mW}(\vx^{(i)})$
    \ENDFOR

    \STATE $\hat{\vy} \leftarrow [\hat{y}_1, \hat{y}_2, ..., \hat{y}_b]$ \COMMENT{{\scriptsize Concatenate all predictions}} \\

    \STATE Compute the training loss $L$:\\\hspace{\algorithmicindent}$L = CrossEntropyLoss(\vy,\hat{\vy}) + \lambda \cdot (s_1 + s_2 + ... + s_D)$ \\ 
    \STATE Compute the gradient of the loss $L$ w.r.t.:\\ \hspace{\algorithmicindent} $\mW, \theta_{\text{WPN}}, \theta_{\text{SPN}}$ using backpropagation\\
    \STATE Update the parameters:\\\hspace{\algorithmicindent} $\mW \leftarrow \mW - \alpha \nabla_{\mW} L,$\\\hspace{\algorithmicindent}$ \theta_{\text{WPN}} \leftarrow \theta_{\text{WPN}} - \alpha \nabla_{\theta_{\text{WPN}}} L,$\\\hspace{\algorithmicindent}$\theta_{\text{SPN}} \leftarrow \theta_{\text{SPN}} - \alpha \nabla_{\theta_{\text{SPN}}} L$

\ENDFOR

\end{algorithmic}

\textbf{Return}: Trained models $f_{\mW}, f_{\theta_{\text{WPN}}}, f_{\theta_{\text{SPN}}}$
\end{algorithm}

\subsection{Training Objective}
\label{sec:training-objective}

The training objective is composed of the average cross-entropy loss and a simple sparsity loss. For each sample, the cross-entropy loss $\ell \left( f_{\mW}(\vx^{(i)}), y_i ; f_{\vtheta_{\text{WPN}}}, f_{\vtheta_{\text{SPN}}} \right)$ is computed between the label predicted using the classification network $f_{\mW}(\vx^{(i)})$ and the true label $y_i$. Note that the auxiliary networks $f_{\vtheta_{\text{WPN}}}$ and $f_{\vtheta_{\text{SPN}}}$ are used to compute the weights of the first layer of the classifier $f_{\mW}$ and this enables computing gradients for their parameters ($\vtheta_{\text{WPN}}, \vtheta_{\text{SPN}}$).

We define a differentiable \textbf{sparsity loss} to incentivise the classifier to use a small number of features. The sparsity loss is the sum of the SPN's feature importance scores $s_j$ output. Note that $s_j > 0$ because they are output from a sigmoid function. To provide intuition for our sparsity loss, we remark that it is a special case of the $L_1$ norm with all terms being strictly positive. Adding $L_1$ regularisation is effective in performing feature selection \citep{tibshirani1996lasso}.

\begin{dmath}
L(\mW, \vtheta_{\text{WPN}}, \vtheta_{\text{SPN}}) = \frac{1}{N} \sum_{i=1}^N \ell \left( f_{\mW}(\vx \esv{i}), y_i; f_{\vtheta_{\text{WPN}}}, f_{\vtheta_{\text{SPN}}} \right) + \lambda \sum_{j=1}^D f_{\vtheta_{\text{SPN}}}(\ve^{(j)})
\end{dmath}

\paragraph{Optimisation.} As the loss is differentiable, we compute the gradients for the parameters of all three networks and use standard gradient-based optimisation algorithms (e.g., AdamW). The three networks are trained simultaneously end-to-end to minimise the objective function. We did not observe optimisation issues by training over $10000$ models.

\section{Experiments}
\label{sec:experiments}

In this section, we evaluate our method for classification tasks and substantiate the proposed architectural choices. Through a series of experiments, we analyse multiple feature embedding types (Sec. \ref{sec:feature-embedding-performance}), the impact of the auxiliary Sparsity Network on the overall performance of WPFS, and its ability to perform feature selection (Sec. {sec:influence-spn-performance}). We then compare the performance of WPFS against the performance of several common methods (Sec. \ref{sec:predictive-performance}) and analyse the training behaviour of WPFS (Sec.~ \ref{sec:training-behaviour}). Lastly, we showcase how the SPN can provide insights into the learning task (Sec. \ref{sec:spn-interpretability}).

\paragraph{Datasets.} We focus on high-dimensional, small size datasets and consider $9$ real-world tabular biomedical datasets with continuous values. The datasets contain $3312-19993$ features, and between $100-200$ samples. The number of classes ranges between $2-4$. Appx. \ref{appendix:datasets} presents complete details about the datasets.

\paragraph{Setup and evaluation.} For each dataset, we perform $5$-fold cross-validation repeated $5$ times, resulting in $25$ runs per model. We randomly select $10\%$ of the training data for validation for each fold. We purposely constrain the size of the datasets (even for larger ones) to simulate real-world practical scenarios with small size datasets. For training all methods, we use a weighted loss (e.g., weighted cross-entropy loss for all neural networks). For each method, we perform a hyper-parameter search and choose the optimal hyper-parameters based on the performance of the validation set. We perform model selection using the validation cross-entropy loss. Finally, we report the balanced accuracy of the test set, averaged across all $25$ runs. Appx. \ref{appendix:reprodicibility} contains details on reproducibility and hyper-parameter tuning.

\paragraph{WPFS settings.} The classification network is a $3$-layer feed-forward neural network with $100, 100, 10$ neurons and a softmax activation in the last layer. The auxiliary WPN and SPN networks are $4$-layer feed-forward neural networks with $100$ neurons. WPN has a $tanh$ activation in the last layer, and SPN has a $sigmoid$ activation. All networks use batch normalisation \citep{ioffe2015batch}, dropout \citep{srivastava2014dropout} with $p=0.2$ and LeakyReLU nonlinearity internally. We train with a batch size of $8$ and optimise using AdamW \citep{Loshchilov2019DecoupledWD} with a learning rate $3e-3$ and weight decay of $1e-4$. We use a learning-rate scheduler, linearly decaying the learning rate from $3e-3$ to $3e-4$ over $500$ epochs, and continue training with $3e-4$ \footnote{We obtained similar results by training with constant learning rate of $1e-4$. Using the cosine annealing with warm restarts scheduler \citep{Loshchilov2017SGDRSG} led to unstable training.}.

\begin{table*}[t!]
\centering
\small
\begin{tabular}{lccccccccccc}
\toprule
Dataset      & \textbf{WPFS (ours)} & DietNets & FsNet & CAE  & LassoNet & DNP  & SPINN & TabNet & MLP  & RF & LGBM \\ \midrule
cll              & 79.14 & 68.84 & 66.38 & 71.94 & 30.63 & 85.12 & 85.34 & 57.81 & 78.30 & 82.06 & 85.59 \\
lung             & 94.83 & 90.43 & 91.75 & 85.00 & 25.11 & 92.83 & 92.26 & 77.65 & 94.20 & 91.81 & 93.42 \\
metabric-dr      & 59.05 & 56.98 & 56.92 & 57.35 & 48.88 & 55.79 & 56.13 & 49.18 & 59.56 & 51.38 & 58.23 \\
metabric-pam50   & 95.96 & 95.02 & 83.86 & 95.78 & 48.41 & 93.56 & 93.56 & 83.60 & 94.31 & 89.11 & 94.97 \\
prostate         & 89.15 & 81.71 & 84.74 & 87.60 & 54.78 & 90.25 & 89.27 & 65.66 & 88.76 & 90.78 & 91.38 \\
smk              & 66.89 & 62.71 & 56.27 & 59.96 & 51.04 & 66.89 & 68.43 & 54.57 & 64.42 & 68.16 & 65.70 \\
tcga-2ysurvival  & 59.54 & 53.62 & 53.83 & 59.54 & 46.08 & 58.13 & 57.70 & 51.58 & 56.28 & 61.30 & 57.08 \\
tcga-tumor-grade & 55.91 & 46.69 & 45.94 & 40.69 & 33.49 & 44.71 & 44.28 & 39.34 & 48.19 & 50.93 & 49.11 \\
toxicity         & 88.29 & 82.13 & 60.26 & 60.36 & 26.67 & 93.5  & 93.5  & 40.06 & 93.21 & 80.75 & 82.40 \\ \midrule
Average rank & \textbf{2.66}        & 6.66     & 7.88  & 6.22 & 11       & 4.33 & 4.33  & 10     & 4.44 & 4.55          & 3.44 \\ \bottomrule
\end{tabular}

\caption{Evaluation of WPFS and $8$ baselines methods on $9$ biomedical datasets. We report the balanced accuracy averaged over $25$ runs. On each dataset, we rank the methods based on their performance. We compute the average rank of each method across datasets (a smaller rank implies higher performance). WPFS ranks the best across all methods.}
\label{tab:main-results}
\end{table*}

\begin{figure}[t!]
    \centering
    \includegraphics[width=\linewidth]{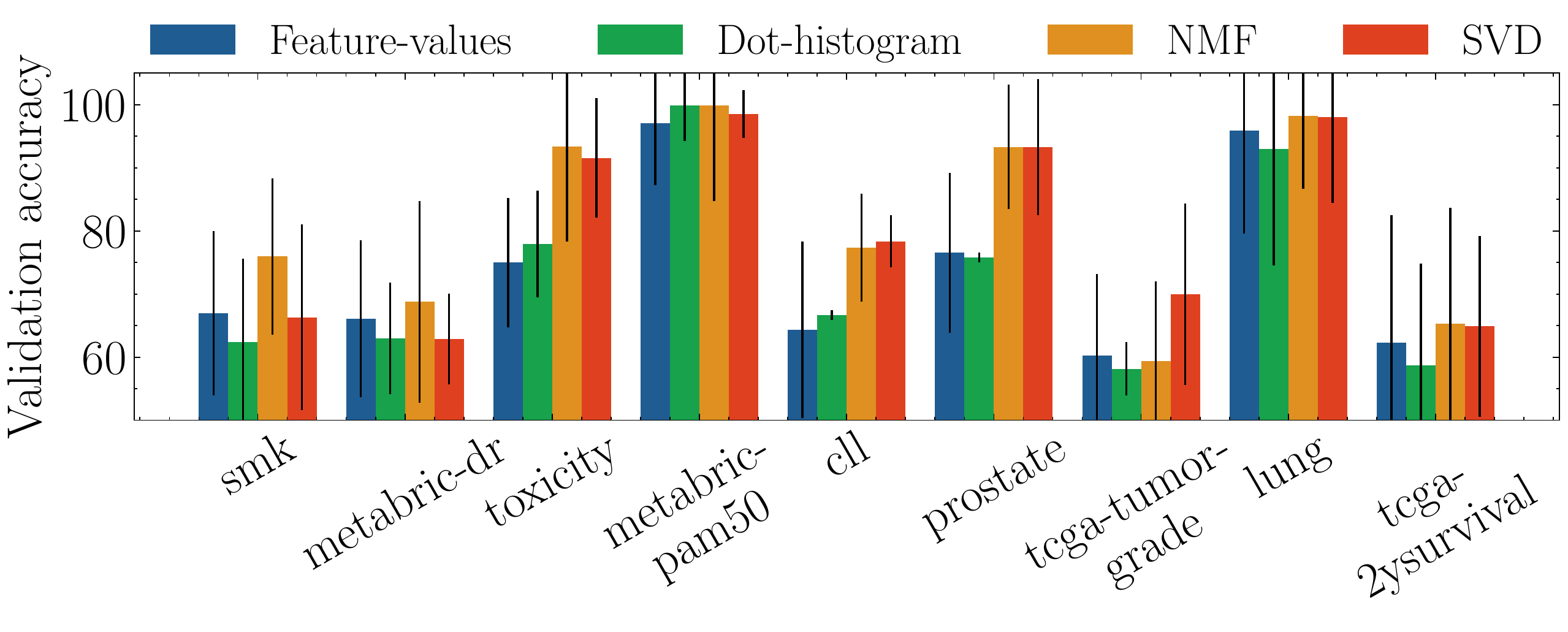}
    \caption{Comparing the validation performance of WPFS with four feature embedding types. We report the mean and standard deviation of the balanced accuracy, averaged across data splits. Across $9$ datasets, feature embeddings based on matrix factorisation (SVD and NMF) outperform `feature-values' and `dot-histogram' embeddings.}
    \label{fig:feature_embedding}
\end{figure}

\subsection{Impact of the Feature Embedding Type}
\label{sec:feature-embedding-performance}

We start by investigating the impact of the feature embedding type on the model's performance. We use and report the validation performance (rather than test performance) not to overfit the test data. The two auxiliary networks take as input feature embeddings to compute the weights of the classifier's first layer $\mW^{[1]}$. Different approaches exist for preprocessing the data $\mX$ before computing the embeddings. We found that $[0, 1]$ scaling leads to better and more stable results than using the raw data or performing Z-score normalisation (Appx. \ref{appendix:embeddings_comparison} presents extended results).

We evaluate the classification performance using four embedding types. In this experiment, we fix the size of the embedding to $50$ and remove the SPN network so that $\mW^{[1]} = \mW_{\text{WPN}}$. In principle, dropping the SPN and using the dot-histogram is a close adaptation of DietNetworks to continuous data, and it allows us to evaluate our statements in Section \ref{sec:rw} that the embedding type affects the performance.

In general, we found that using NMF embeddings leads to more stable and accurate performance (Figure~\ref{fig:feature_embedding}). In all except one case, the performance of the NMF embeddings is better or on-par with the SVD embeddings, with both exhibiting substantially better performance than dot-histogram and feature-values. In most cases, the ``simple'' feature-values embedding outperforms the dot-histogram embeddings. We believe dot-histogram cannot represent continuous data accurately because the choice of bins can be highly impacted by the outliers and the noise in the data. We use NMF embeddings in all subsequent sections.

We also investigated the influence of the size of the embeddings on the performance. We found that in most cases, an embedding of size $50$ led to optimal results, with optimal sizes for `lung' and `cll' being $20$ and $70$, respectively (Appx.~\ref{appendix:embeddings_comparison} contains extended results of this analysis).

\subsection{Impact of the Feature Selection Mechanisms}
\label{sec:influence-spn-performance}

\begin{figure}[t!]
    \centering
    \includegraphics[ width=0.9\linewidth]{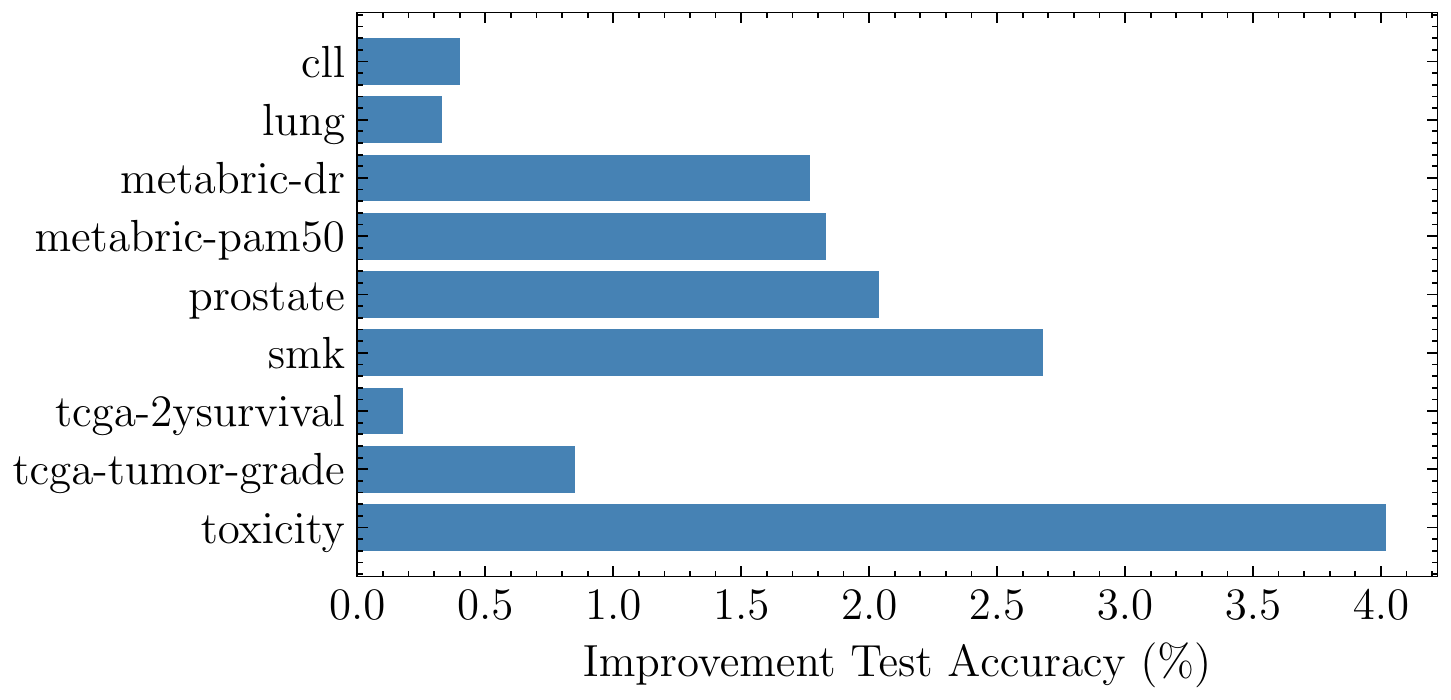}
    \caption{The effect of SPN on the WPFS performance. We report the average test accuracy improvement of a WPFS which includes a SPN over an identical WPFS without SPN. We trained without the sparsity loss ($\lambda = 0$) to isolate SPN's effect. Adding the SPN always improves the performance.}
    \label{fig:ablation_spn}
\end{figure}

Next, we analyse how the \textbf{auxiliary SPN} impacts the overall performance. To isolate the effect of the SPN, we remove the sparsity loss (i.e., set $\lambda=0$) and train and evaluate two identical setups: one with and one without the auxiliary SPN. Figure \ref{fig:ablation_spn} shows the improvement in test accuracy (Appx. \ref{appendix:ablation-spn} presents extended results). In all cases, adding the SPN network leads to better performance. While these improvements vary from one dataset to another, the results show that the SPN is an appropriate regularisation mechanism and reduces potential unwanted saturations of the WPN \footnote{Recall that the last activation of the WPN is a $tanh$, which can saturate during training and output weights with values $\in \{-1, 1\}$.}.

Recall that WPFS can be trained using an \textbf{additional sparsity loss} computed from the output $\vs$ of the SPN. We study the effect of the sparsity loss by training the complete WPFS model (including the WPN and SPN) and varying the hyper-parameter $\lambda$ which controls the magnitude of the sparsity loss. We observe that the sparsity loss increases the performance of WPFS on $8$ out of $9$ datasets (Appx. \ref{tab:varying-lambda-performance} contains the ablation on $\lambda$). In general, $\lambda= 3e-5$ leads to the best performance improvement, but choosing $\lambda$ is still data-dependent and it should be tuned. We suggest trying values for $\lambda$ such that the ratio between the cross-entropy loss and the sparsity loss is $\{ 0.05, 0.1, 0.2, 0.5, 1\}$.

\subsection{Classification Performance of WPFS}
\label{sec:predictive-performance}
Having investigated the individual impact of the proposed architectural choices, we proceed to evaluate the test performance of WPFS against a version of DietNetworks adapted for continuous data by using the dot-histogram embeddings presented in Section \ref{sec:feature-embedding-description}. We compare to methods mentioned in Related Works, including FsNet, CAE, LassoNet\footnote{We discuss LassoNet training instabilities in Appx. \ref{appendix:hyperparam}}, SPINN and DNP. We also compare to standard methods such as TabNet \citep{arik2021tabnet}, Random Forest (RF) \citep{breiman2001random}, LightGBM (LGBM) \citep{ke2017lightgbm} and an MLP. We report the mean balanced test accuracy, and include the standard deviations in Appx. \ref{sec:full-results-additional-baselines}.

WPFS exhibits better or comparable performance than the other benchmark models (Table \ref{tab:main-results}). Specifically, we computed the average rank of each model's performance across tasks and found that WPFS ranks best, followed by tree-based ensembles (LightGBM and Random Forest) and neural networks performing feature sparsity using $L_1$ (SPINN, DNP). Note that WPFS consistently outperformed other `parameter-efficient' models such as DietNetworks and FsNet. We also found that complex and parameter-heavy architectures such as TabNet, CAE and LassoNet struggle in these scenarios and are surpassed by simple but well-regularised MLPs, which aligns with the literature \citep{kadra2021well}.

In most cases, WPFS is able to perform substantially better than MLPs, struggling only in the case of `toxicity'. As the MLP is already well-regularised, we attribute the performance improvement of WPFS to its ability to reduce the number of learnable parameters and therefore overcome severe overfitting, typical for such small size data problems.

\subsection{Training Behaviour}
\label{sec:training-behaviour}

\begin{figure}[t!]
    \centering
    \includegraphics[width=\linewidth]{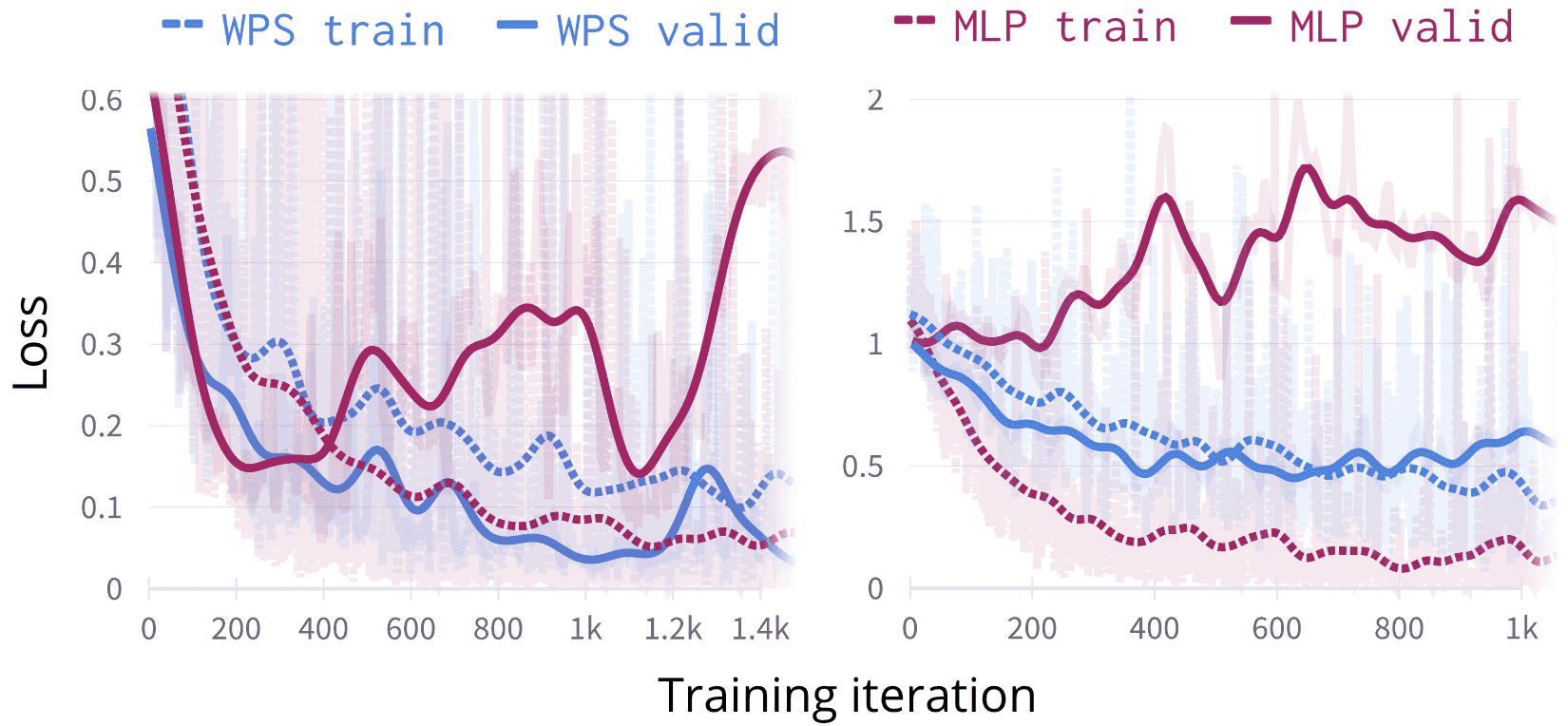}
    \captionof{figure}{Train and validation loss curves for an MLP and WPFS with an identical architecture for the classification network, on `lung' and `tcga-tumor-grade' datasets. The MLP shows high overfitting, as the validation loss quickly diverges while the training loss converges. WPFS's validation loss decreases for longer and achieves better minima.}
    
    \label{fig:overfitting_trends}
\end{figure}

Since WPFSs are able to outperform other neural networks, we attribute this to the additional regularisation capabilities of the WPN and SPN networks. To investigate this hypothesis, we analysed the trends of the train/validation loss curves of WPFS (with WPN and SPN) and MLP. Figure \ref{fig:overfitting_trends} shows the train and validation loss curves for training on `lung' and `tcga-tumor-grade' datasets (see Appx. \ref{appendix:training_dynamics} for extended results for all datasets). Although the training loss consistently decreases for both models, there is a pronounced difference in the validation loss convergence. Specifically, the validation loss of the MLP diverges quickly and substantially from the training loss, indicating overfitting. In contrast, the WPFS validation loss indicates less overfitting because it decreases for longer and achieves a lower minimum than the MLP. The disparities in convergence translate into the test results from Table \ref{tab:main-results}, where WPFS outperforms the MLP. We also found that using the WPN in our model leads to performance improvment in all but two cases (‘cll’ and ‘toxicity’), when its addition renders subpar but comparable performance to an MLP (see Appx. \ref{appendix:ablation-wpn}).

WPFS exhibit stable performance, as indicated by the average standard deviation of the test accuracy across datasets (Appx. \ref{sec:full-results-additional-baselines}). This is consistent when varying the sample size, with WPFSs outperforming MLPs on smaller datasets, with the latter closing the performance gap as sample size increases (see Appx. \ref{appendix:ablation-dataset-size}).

Note, however, that these results are obtained using $10\%$ of the training data for validation, which translates into only a handful of samples for such small size datasets.\footnote{The outcome was similar with $15\%$ validation splits.} While having small validation sets is not ideal and greatly influences model selection (evident from the instabilities in the validation loss curves), we purposely constrained the experimental setting to mimic a real-world practical scenario. Therefore, obtaining stable validation loss and performing model selection remains an open challenge.

\subsection{Feature Importance and Interpretability}
\label{sec:spn-interpretability}

\begin{figure}[t!]
    \centering
    \includegraphics[width=\linewidth]{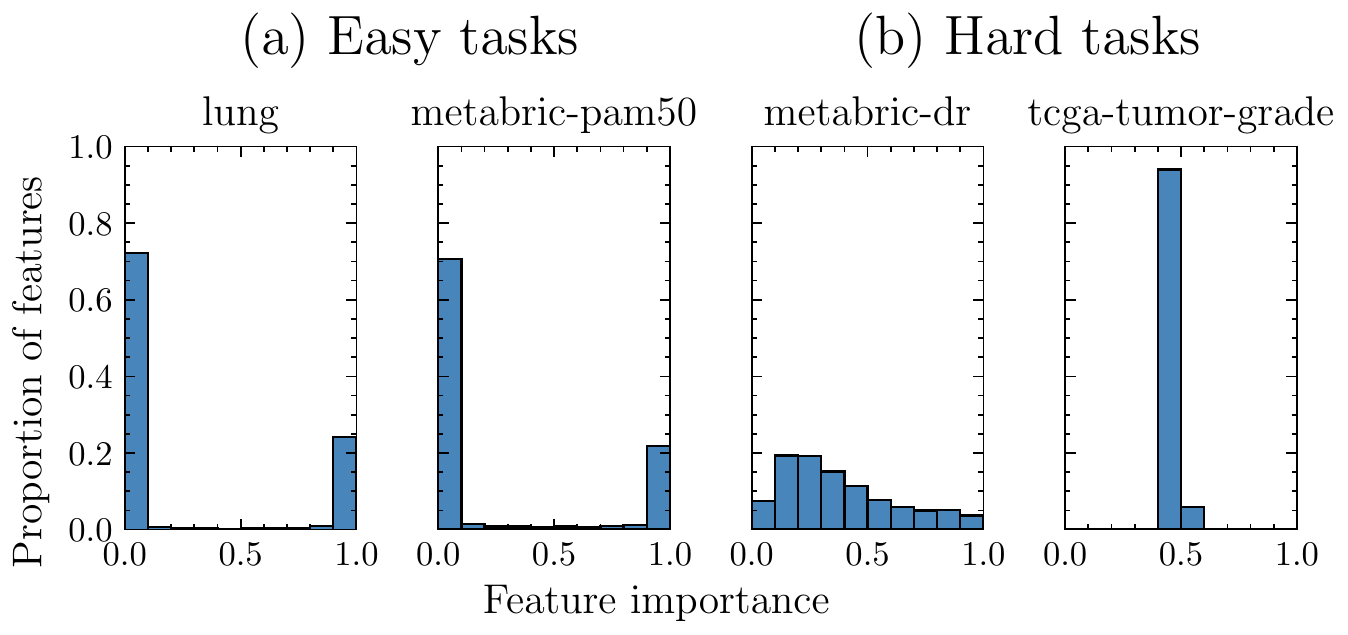}
    \caption{The distribution of the feature importance scores computed by WPFS provides insights into the difficulty of the learning task. WPFS assigns almost binary feature importance scores for `easy' tasks while it struggles to make a `clear-cut' of the important features on `hard' tasks.}
    \label{fig:feature_selection_distribution_easy_hard_tasks}
\end{figure}

In Section \ref{sec:influence-spn-performance} we showed that the proposed feature selection mechanism consistently improves the performance. In our final set of analyses, we seek qualitative insights into the feature selection mechanism.

\begin{figure}
    \centering
    \includegraphics[width=0.90\columnwidth]{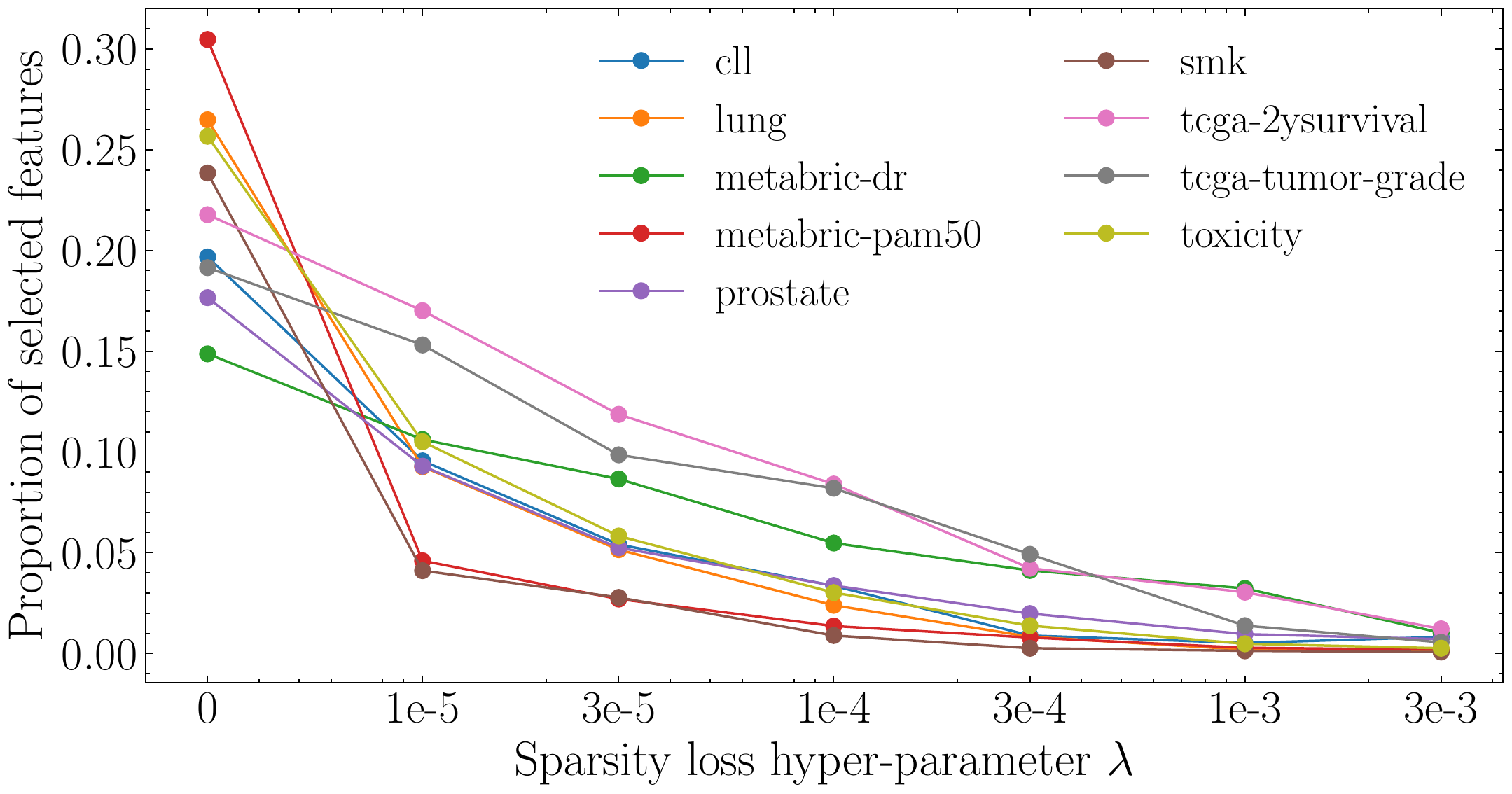}
    
    \caption{Proportion of selected features for different sparsity loss hyper-parameter $\lambda$, averaged across all runs. Increasing $\lambda$ reduces the proportion of selected features from $\sim 25\%$ to less than $1\%$, enabling post-hoc analyses of the selected features, possibly using domain knowledge.}
    \label{fig:increasing_lambda_proportion_selected_features}
\end{figure}

Firstly, we examine how \textbf{varying the sparsity loss} impacts the WPFS's ability to determine the important features. Let's consider a feature $j$ is being selected when the score $s_j > 0.95$, and not selected when $s_j < 0.95$. Figure \ref{fig:increasing_lambda_proportion_selected_features} shows the proportion of selected features while varying the sparsity hyper-parameter $\lambda$. Notice that, across datasets, WPFS selects merely $\sim 25\%$ of features without being trained with the sparsity loss (i.e., $\lambda = 0$), which we believe is due to using a sigmoid activation for SPN. By increasing $\lambda$, the proportion of selection features decreases to $\sim 0.1 - 1\%$. In Appx. \ref{appendix:feature_importance_distribution_varying_lambda}) we show that increasing $\lambda$ changes the \textit{distribution} of feature importance scores in a predictable manner. Although we expected these outcomes, we believe they are essential properties of WPFS, which can enable interpretable post-hoc analyses of the results.

Lastly, we discovered that the \textbf{feature importance scores} $\vs$ computed by WPFS give insight into the task's difficulty. We denote a task as `easy' when WPFS obtains high accuracy and `hard' when WPFS obtains low accuracy. In this experiment, we analysed the performance of WPFS without sparsity loss (by setting $\lambda = 0$). On easy tasks (Fig. \ref{fig:feature_selection_distribution_easy_hard_tasks} (a)), we observe that WPFS has high certainty and assigns almost binary $\in \{ 0, 1 \}$ feature importance scores, with a negligible proportion of features in between. In contrast, on hard tasks (Fig. \ref{fig:feature_selection_distribution_easy_hard_tasks} (b)), the model behaves conservatively, and it cannot rank the features with certainty (this is most evident on `tcga-tumor-grade', where WPFS assigns $0.5$ feature importance score to all features). This behaviour provides a built-in mechanism for debugging training issues, and analysing it from a domain-knowledge perspective may lead to further performance improvements. We conjecture that these findings link to WPFS's ability to identify spurious correlations, and we leave this investigation for future work.

\section{Conclusion}
We present \textbf{W}eight \textbf{P}redictor Network with \textbf{F}eature \textbf{S}election (WPFS), a general method for learning from high-dimensional but extremely small size tabular data. Although well-regularised simple neural networks outperform more sophisticated architectures on tabular data, they are still prone to overfitting on very small datasets with many potentially irrelevant features. Motivated by this, we design an end-to-end method that combines a feed-forward network with two auxiliary networks: a Weight Predictor Network (WPN) that reduces the number of learnable parameters and a Sparsity Network (SPN) that performs feature selection and serves as an additional regularisation mechanism. 

We investigated the capabilities of WPFS through a series of experiments on nine real-world biomedical datasets. We demonstrated that WPFS outperforms ten methods for tabular data while delivering the most stable classification accuracy. These results are encouraging, especially because neural networks typically struggle in this setting of high-dimensional, small size tabular datasets. We attribute the success of WPFS to its ability to reduce overfitting. We analysed the train-validation loss curves and showed that WPFS's validation loss converges for longer and achieves better minima than an MLP. Lastly, we analysed the feature selection mechanism of WPFS and uncovered favourable properties, such as using only a small set of features that can be analysed, possibly using domain knowledge.

\section*{Acknowledgments}
We would like to thank the reviewers for their feedback and efforts towards improving our paper. We acknowledge the support from the Cambridge ESRC Doctoral Training Partnership for A.M., and the support from The Mark Foundation for Cancer Research and Cancer Research UK Cambridge Centre [C9685/A25177] for N.S., P.L. and M.J.

\bibliography{aaai23.bib}

\clearpage
\appendix
\onecolumn

\section*{Supplementary Material for Submission ``Weight Predictor Network with Feature Selection for Small Sample Tabular Biomedical Data"}

\section{Reproducibility}
\label{appendix:reprodicibility}

\subsection{Datasets} 
\label{appendix:datasets}

\begin{table*}[h!]
  \centering
  \caption{Details of the 9 real-world biomedical datasets used for experiments. The number of features is 15-110 times larger than the number of samples.}
    \label{tab:datasets}
  
    \begin{tabular}{@{}lcccr@{}}
    \toprule
    \textbf{Dataset} & \textbf{\# Samples} & \textbf{\# Features} & \textbf{\# Classes} & \textbf{\# Samples per class} \\
    \midrule
    cll & 111   & 11340 & 3     & 11, 49, 51 \\
    lung  & 197   & 3312  & 4    & 17, 20, 21, 139  \\
    metabric-dr & 200   & 4160  & 2     & 61, 139 \\
    metabric-pam50 & 200   & 4160  & 2     & 33, 167 \\
    prostate & 102   & 5966  & 2     & 50, 52 \\
    smk & 187   & 19993 & 2     & 90, 97 \\
    tcga-2ysurvival & 200   & 4381  & 2     & 78, 122 \\
    tcga-tumor-grade & 200   & 4381  & 3     & 25, 52, 124 \\
    toxicity & 171   & 5748  & 4     & 39, 42, 45, 45 \\ \bottomrule
    \end{tabular}%
\end{table*}

Table \ref{tab:datasets} summarises the datasets. Five datasets are open-source \cite{li2018feature} and available online (https://jundongl.github.io/scikit-feature/datasets.html): \textbf{CLL-SUB-111} (called `cll') \citep{haslinger2004microarray}, \textbf{lung} \citep{bhattacharjee2001classification}, \textbf{Prostate\_GE} (called `prostate') \citep{singh2002gene}, \textbf{SMK-CAN-187} (called `smk') \citep{spira2007airway} and \textbf{TOX-171} (called `toxicity') \citep{bajwa2016cutting}.

We derived two datasets from the \textbf{METABRIC} \citep{curtis2012genomic} dataset. We combined the molecular data with the clinical label `DR' to create the \textbf{`metabric-dr'} dataset, and we combined the molecular data with the clinical label `Pam50Subtype' to create the \textbf{`metabric-pam50'} dataset. Because the label `Pam50Subtype' was very imbalanced, we transformed the task into a binary task of basal vs non-basal by combining the classes `LumA', `LumB', `Her2', `Normal' into one class and using the remaining class `Basal' as the second class. For both `metabric-dr' and `metabric-pam50' we selected the Hallmark gene set \citep{liberzon2015molecular} associated with breast cancer, and the new datasets contain $4160$ expressions (features) for each patient. We created the final datasets by randomly sampling $200$ patients stratified because we are interested in studying datasets with a small sample size. 

We derived two datasets from the \textbf{TCGA} \citep{tomczak2015cancer} dataset. We combined the molecular data and the label `X2yr.RF.Surv' to create the \textbf{`tcga-2ysurvival'} dataset, and we combined the molecular data and the label `tumor\_grade' to create the \textbf{`tcga-tumor-grade'} dataset. For both `tcga-2ysurvival' and `tcga-tumor-grade' we selected the Hallmark gene set \citep{liberzon2015molecular} associated with breast cancer, leaving $4381$ expressions (features) for each patient. We created the final datasets by randomly sampling $200$ patients stratified because we are interested in studying datasets with a small sample size.

\paragraph{Dataset processing} Before training the models, we apply Z-score normalisation to each dataset. Specifically, on the training split, we learn a simple transformation to make each column of $\mX_{train} \in \sR^{N_{train} \times D}$ have zero mean and unit variance. We apply this transformation to the validation and test splits during cross-validation.

\subsection{Computing Resources}
\label{appendix:computing-resources}

We trained over $40,000$ models on a single machine from an internal cluster with a GPU Nvidia Quadro RTX 8000 with 48GB memory and an Intel(R) Xeon(R) Gold 5218 CPU @ 2.30GHz with 16 cores. The operating system was Ubuntu 20.04.4 LTS.

\subsection{Training Details and Hyper-parameter Tuning}
\label{appendix:hyperparam}

\paragraph{Software implementation.} All software dependencies, alongside their versions, are specified in the associated code. We implemented Random Forest using scikit-learn \citep{pedregosa2011scikit}, LightGBM using the lightgbm library \citep{ke2017lightgbm} and TabNet \citep{arik2021tabnet} using a popular implementation from Dreamquark AI (www.github.com/dreamquark-ai/tabnet). We re-implemented FsNet \citep{singh2020fsnet} because the official code implementation contains differences from the paper, and they used a different evaluation setup from ours (they evaluated using unbalanced accuracy, while we run multiple data splits and evaluate using balanced accuracy). We used the official implementation of LassoNet (https://github.com/lasso-net/lassonet), and the official implementation of SPINN (https://github.com/jjfeng/spinn). We implemented MLP, DietNetworks, CAE our proposed WPFS in PyTorch \citep{paszke2019pytorch}. We implemented NMF using a popular GPU-based NMF package (www.github.com/yoyololicon/pytorch-NMF).

\paragraph{Training details.} We present the most important training settings (all other implementation details are included in the associated codebase). The per-dataset hyper-parameters are discussed in the next paragraph and Table \ref{table:best-hyper-params}. We perform a fair comparison whenever possible. For example, use train all models using a weighted loss, evaluate using balanced accuracy, and use the same classification network architecture for WPFS, MLP, FsNet, CAE and DietNetworks.

\begin{itemize}
    \item \textbf{WPFS, DietNetworks, FsNet, CAE} have three hidden layers of size $100, 100, 10$. The Weight Predictor Network and the Sparsity Network have four hidden layers $100, 100, 100, 100$. They are trained for $10000$ iterations using early stopping with patience $200$ on the validation cross-entropy and gradient clipping at $2.5$. For \textbf{CAE} and \textbf{FsNet} we use the suggested annealing schedule for the concrete nodes: exponential annealing from temperature $10$ to $0.01$.
    On all datasets, \textbf{DietNetworks} performed best with not decoder.
    \item \textbf{LassoNet} has three hidden layers of size $100, 100, 10$. We use dropout $0.2$, and train using AdamW (with betas $0.9, 0.98$) and a batch size of 8. We train using a weighted loss. We perform early stopping on the validation set.
    \item For \textbf{Random Forest} we used $500$ estimators, feature bagging with the square root of the number of features, and used balanced weights associated with the classes. 
    \item For \textbf{LightGBM} we used $200$ estimators, feature bagging with $30\%$ of the features, a minimum of two instances in a leaf, and trained for $10000$ iterations to minimise the balanced cross-entropy. We perform early stopping on the validation set.
    \item For \textbf{TabNet} we use width 8 for the decision prediction layer and the attention embedding for each mask (larger values lead to severe overfitting), and $1.5$ for the coefficient for feature reusage in the masks. We use three steps in the architecture, with two independent and two shared Gated Linear Units layers at each step. We train using Adam with momentum $0.3$ and clip the gradients at $2$. 
    \item For \textbf{SPINN} we used $\alpha=0.9$ for the group lasso in the sparse group lasso, the sparse group lasso hyper-parameter $\lambda=0.0032$, ridge-param $\lambda_0 = 0.0001$, and train for at most $1000$ iterations.
    \item For \textbf{DNP} we did not find any suitable implementation, and used SPINN with different settings as a proxy for DNP (because DNP is a greedy approximation to optimizing the group lasso, and SPINN optimises directly a group lasso). Specifically, our proxy for DNP results is SPINN with $\alpha=1$ for the group lasso in the sparse group lasso, the sparse group lasso hyper-parameter $\lambda=0.0032$, ridge-param $\lambda_0 = 0.0001$, and train for at most $1000$ iterations.
\end{itemize}

\begin{table}[h!]
\centering
\caption{Best performing hyper-parameters for each baseline and each dataset.}
\label{table:best-hyper-params}

\begin{tabular}{@{}lccccccccc@{}}
\toprule
Dataset & \multicolumn{2}{c}{Random Forest} & \multicolumn{2}{c}{LightGBM} & \multicolumn{2}{c}{TabNet}    & \multicolumn{2}{c}{FsNet} & CAE       \\ \midrule
        & max         & min samples         & learning        & max        & learning & $\lambda$ sparsity & learning    & $\lambda$    & annealing \\
                 & depth & leaf & rate & depth & rate &       & rate  & reconstruction & iterations \\ \midrule
cll              & 3     & 3    & 0,1  & 2     & 0,03 & 0,001 & 0,003 & 0              & 1000       \\
lung             & 3     & 2    & 0,1  & 1     & 0,02 & 0,1   & 0,001 & 0              & 1000       \\
metabric-dr      & 7     & 2    & 0,1  & 1     & 0,03 & 0,1   & 0,003 & 0              & 300        \\
metabric-pam50   & 7     & 2    & 0,01 & 2     & 0,02 & 0,001 & 0,003 & 0              & 1000       \\
prostate         & 5     & 2    & 0,1  & 2     & 0,02 & 0,01  & 0,003 & 0              & 1000       \\
smk              & 5     & 2    & 0,1  & 2     & 0,03 & 0,001 & 0,003 & 0              & 1000       \\
tcga-2ysurvival  & 3     & 3    & 0,1  & 1     & 0,02 & 0,01  & 0,003 & 0              & 300        \\
tcga-tumor-grade & 3     & 3    & 0,1  & 1     & 0,02 & 0,01  & 0,003 & 0              & 300        \\
toxicity         & 5     & 3    & 0,1  & 2     & 0,03 & 0,1   & 0,001 & 0,2            & 1000       \\ \bottomrule
\end{tabular}

\end{table}

\paragraph{Hyper-parameter tuning.} For each model, we used random search and previous experience to find a good range of hyper-parameter values that we could investigate in detail. After we found a good range for the hyper-parameters, we performed a grid-search and trained using 5-fold cross-validation with 5 repeats (training 25 models each run). 

For the MLP and DietNetworks we individually grid-searched learning rate $\in \{ 0.003, 0.001, 0.0003, 0.0001 \}$, batch size $ \in \{ 8, 12, 16, 20, 24, 32 \} $, dropout rate $\in \{ 0, 0.1, 0.2, 0.3, 0.4, 0.5 \}$. We found that learning rate $0.003$, batch size $8$ and dropout rate $0.2$ work well across datasets for both models, and we used them in the presented experiments. In addition, for DietNetworks we also tuned the reconstruction hyper-parameter $\lambda \in \{0, 0.01, 0.03, 0.1, 0.3, 1, 3, 10, 30 \}$ and found that across dataset having $\lambda = 0 $ performed best. For WPFS we used the best hyper-parameters for the MLP and tuned only the sparsity hyper-parameter $\lambda \in \{ 0, 3e-6, 3e-5, 3e-4, 3e-3, 1e-2 \}$ (Appx. \ref{appendix:varying-lambda-performance}) and the size of the feature embedding $\in \{ 20, 50, 70\}$ (Appx. \ref{appendix:ablation-feature-embedding-size}). For CAE we grid-search the learning rate in $\{0.0001, 0.001\}$ and the number of annealing iterations in $\{ 300, 1000\}$ (the models train for $\sim 1000-2000$ iterations with early stopping). For FsNet we grid-search the reconstruction parameter in $\lambda \in \{ 0, 0.2, 1, 5 \}$ and learning rate in $\{0.001, 0.003\}$. For LightGBM we performed grid-search for the learning rate in $\{0.1, 0.01\}$ and maximum depth in $\{1, 2\}$. For Random Forest, we performed a grid search for the maximum depth in $\{3, 5, 7 \}$ and the minimum number of samples in a leaf in $\{2, 3\}$. For TabNet we searched the learning rate in $\{0.01, 0.02, 0.03\}$ and the $\lambda$ sparsity hyper-parameter in $\{0.1, 0.01, 0.001, 0.0001\}$, as motivated by \cite{Yang2021LocallySN}. We selected the best hyper-parameters (Table \ref{table:best-hyper-params}) on the weighted cross-entropy (except Random Forest, for which we used weighted balanced accuracy).

\paragraph{LassoNet unstable training.} We used the official implementation of LassoNet (https://github.com/lasso-net/lassonet), and we successfully replicated some of the results in the LassoNet paper \citep{lemhadri2021lassonet}.
However, LassoNet was unstable on all the $9$ datasets we trained on. We included a Jupyter notebook in the attached codebase that demostrates that LassoNet cannot even fit the training data on our datasets.

In our experiments, we grid-searched the $L_1$ penalty coefficient $\lambda \in \{ 0.001, 0.01, 0.1, 1, 10, `auto'\}$ and the hierarchy coefficient $M \in \{ 0.1, 1, 3, 10 \}$. These values are suggested in the paper and used in the examples from the official codebase. For all hyper-parameter combinations, LassoNet's performance was equivalent to a random classifier (e.g., $25\%$ balanced accuracy for a $4$-class problem).

\paragraph{NMF.} We used standard NMF to compute a rank $k$ approximation ($k$ is mentioned in each experiment, but usually $k=50$) of the data matrix $\mX \approx W H$, where $H \in \sR^{k \times D}$. A column $H_{:i}$ represents the embedding for feature $i$. We trained the NMF for 1000 iterations to minimise the error under the Frobenius norm. We computed embeddings only on the training split and used the resulting embeddings also during validation and testing.

\section{Feature Embeddings}
\label{appendix:embeddings_comparison}

\subsection{Ablation Data Preprocessing for Computing the Feature Embeddings}
\label{appendix:embedding-preprocessing}

We investigate the impact of preprocessing the data matrix $\mX$ before computing the embeddings. We consider three types of data preprocessing: \textbf{min-max} rescales the values of each feature in $[0, 1]$, \textbf{Z-score} makes each feature have zero mean, and unit variance and \textbf{raw} does not apply any preprocessing to the data. We fix the size of the embedding to $50$ and remove the effect of the auxiliary SPN network (SPN always outputs $\vs = 1$, such that $\mW^{[1]} = \mW^{[1]}_{WPN}$).

Tables \ref{table-appendix:embedding-preprocessing-1} and \ref{table-appendix:embedding-preprocessing-2} show the validation performance of WPFS on 9 datasets. We consider two types of preprocessing for each embedding type. Z-score normalisation cannot be applied for NMF because NMF requires the input to have all positive values. We find that NMF outperforms `feature-values' and `dot-histogram' by a large margin while being slightly better than SVD. We concluded that min-max preprocessing is most suitable for NMF because it scales the data into a consistent range, reducing the risk that the embeddings have large values. Figure \ref{fig:feature_embedding} is created by plotting the results obtained with min-max preprocessing.

\begin{table*}[h!]
\centering

\caption{Performance of WPFS using `feature-values' and `dot-histogram' feature embeddings. Before computing the embeddings, the data matrix $\mX$ is preprocessed using `min-max' or `Z-score'. We report the mean and standard deviation for the balanced accuracy on the validation split.}
\label{table-appendix:embedding-preprocessing-1}

\begin{tabular}{lrrrr}
\toprule
Embedding type & \multicolumn{2}{c}{\textbf{Feature-values}}    & \multicolumn{2}{c}{\textbf{Dot-histogram}}     \\
Preprocessing  & min-max            & Z-score           & min-max            & Z-score           \\ \midrule
cll            & $64,33 \pm 18,4$  & $62 \pm 15,79$    & $66,67 \pm 16,14$ & $63,67 \pm 15$    \\
lung           & $95,86 \pm 7,2$   & $95,82 \pm 7,28$  & $92,95 \pm 9,45$  & $90,36 \pm 9,67$  \\
metabric-dr    & $66,07 \pm 14$    & $65,56 \pm 13,17$ & $62,98 \pm 12,67$ & $62,87 \pm 14,81$ \\
metabric-pam50   & $97,08 \pm 5,55$  & $96,87 \pm 6,97$  & $99,85 \pm 0,77$  & $96,62 \pm 6,11$  \\
prostate       & $76,56 \pm 15,05$ & $80,8 \pm 17,95$  & $75,76 \pm 15,1$  & $76,8 \pm 11,89$  \\
smk            & $66,96 \pm 13,05$ & $67,54 \pm 10,91$ & $62,43 \pm 12,45$ & $65,18 \pm 10,91$ \\
tcga-2ysurvival  & $62,33 \pm 10,77$ & $62,87 \pm 11,47$ & $58,67 \pm 14,37$ & $63,6 \pm 13,6$   \\
tcga-tumor-grade & $60,27 \pm 12,59$ & $63,53 \pm 11,93$ & $58,13 \pm 11,58$ & $59,47 \pm 13,55$ \\
toxicity       & $75 \pm 20,21$    & $84,46 \pm 11,38$ & $77,92 \pm 13,15$ & $83,75 \pm 13,77$ \\ \midrule

\textbf{Average} & $73,82 \pm 12,98$ & $75,49 \pm 11,87$ & $72,81 \pm 11,74$ & $73,59 \pm 12,14$ \\ \bottomrule
\end{tabular}%

\end{table*}

\begin{table*}[h!]
\caption{Performance of WPFS using Non-negative matrix factorisation (NMF) and Singular Value Decomposition (SVD) feature embeddings. Before computing the embeddings, the data matrix $\mX$ is either not pre-processed (column `raw') or preprocessed using `min-max', `Z-score'. We report the mean and standard deviation for the balanced accuracy on the validation split.}
\label{table-appendix:embedding-preprocessing-2}

\centering
\begin{tabular}{lrrrr}
\toprule
Embedding type & \multicolumn{2}{c}{\textbf{NMF}}               & \multicolumn{2}{c}{\textbf{SVD}}               \\
Preprocessing  & min-max            & raw               & min-max            & Z-score          \\ \midrule
cll            & $77,33 \pm 12,39$ & $80,33 \pm 17,98$ & $78,33 \pm 15,96$ & $77,67 \pm 10,95$ \\
lung           & $98,27 \pm 3,8$   & $97,36 \pm 5,05$  & $98,05 \pm 4,13$  & $97,77 \pm 4,66$  \\
metabric-dr      & $68,76 \pm 12,93$ & $67,67 \pm 12,03$ & $62,87 \pm 16,29$ & $65,89 \pm 15,06$ \\
metabric-pam50   & $99,85 \pm 0,77$  & $99,54 \pm 1,28$  & $98,56 \pm 4,22$  & $98,87 \pm 3,48$  \\
prostate       & $93,3 \pm 8,56$   & $92 \pm 9,19$     & $93,3 \pm 9,86$   & $95,2 \pm 7,21$   \\
smk            & $75,96 \pm 10,23$ & $74,75 \pm 11,59$ & $66,32 \pm 9,8$   & $67 \pm 10,65$    \\
tcga-2ysurvival  & $65,33 \pm 13,58$ & $60,27 \pm 11,06$ & $64,87 \pm 14,31$ & $63,13 \pm 10,47$ \\
tcga-tumor-grade & $59,4 \pm 18,36$  & $65 \pm 14,38$    & $69,93 \pm 14,68$ & $66,07 \pm 12,71$ \\
toxicity       & $93,33 \pm 8,82$  & $93,83 \pm 7,52$  & $91,58 \pm 8,46$  & $93,42 \pm 9,37$  \\ \midrule
\textbf{Average} & $81,28 \pm 9,93$ & $81,19 \pm 10,00$ & $80,42 \pm 10,85$ & $80,55 \pm 9,39$ \\ \bottomrule

\end{tabular}%
\end{table*}

\subsection{Ablation Feature Embedding Size}
\label{appendix:ablation-feature-embedding-size}

We investigate the robustness of WPFS to the size of the embedding. We train the WPFS model (with the two auxiliary networks) following the protocol presented in Section \ref{sec:experiments}, but use a batch size of 16. We use NMF embeddings with min-max preprocessing and set the sparsity hyper-parameter $\lambda = 0$.

Table \ref{table-appendix:embedding-size} shows the \textit{validation} weighted cross-entropy averaged across 25 runs. Varying the embedding size changes little the validation loss, and an embedding of size $50$ performs well across datasets. However, on some datasets WPFS performs better for different embedding sizes (e.g., on `lung' dataset performs substantially better with a smaller embedding). We conclude that WPFS is robust to the choice of the embedding size, and, from a practical perspective, we suggest using an embedding of size $50$ in the first instance.

\begin{table*}[t!]
\centering

\caption{Performance on validation set for different size of the embedding on 9 datasets. We report the mean weighted cross-entropy and standard deviation on the validation set.}
\label{table-appendix:embedding-size}

\begin{tabular}{@{}lrrr@{}}
\toprule
Embedding size   & 20                    & 50                    & 70                    \\ \midrule
cll              & $0,333 \pm 0,21674$   & $0,30634 \pm 0,22093$ & $\mathbf{0,29276 \pm 0,26171}$ \\
lung             & $\mathbf{0,03115 \pm 0,06694}$ & $0,04101 \pm 0,09264$ & $0,04505 \pm 0,08597$ \\
metabric-dr      & $0,53875 \pm 0,11515$ & $\mathbf{0,52352 \pm 0,1422}$  & $0,54482 \pm 0,15124$ \\
metabric-pam50   & $0,00334 \pm 0,01033$ & $\mathbf{0,00032 \pm 0,00092}$ & $0,00688 \pm 0,02191$ \\
prostate         & $0,11005 \pm 0,1617$  & $\mathbf{0,07709 \pm 0,14565}$ & $0,07739 \pm 0,13387$ \\
smk              & $0,45892 \pm 0,14225$ & $\mathbf{0,42537 \pm 0,14262}$ & $0,44695 \pm 0,16388$ \\
tcga-2ysurvival  & $\mathbf{0,51711 \pm 0,15583}$ & $0,53281 \pm 0,16372$ & $0,56536 \pm 0,11952$ \\
tcga-tumor-grade & $0,79771 \pm 0,19892$ & $\mathbf{0,76389 \pm 0,20208}$ & $0,78329 \pm 0,19533$ \\
toxicity         & $0,1509 \pm 0,17211$  & $\mathbf{0,10472 \pm 0,18221}$ & $0,13863 \pm 0,21155$ \\ \midrule
Average          & $0,3267 \pm	0,1377$	& $0,3083 \pm 0,1436$ & $0,3223\pm 0,1494$ \\ \bottomrule

\end{tabular}%

\end{table*}

\clearpage
\section{Ablation Sparsity Network on the Classification Accuracy}
\label{appendix:ablation-spn}

\begin{table}[h!]
\caption{
Test performance of WPFS with and without the auxiliary Sparsity Network (SPN). The experiment setup is presented in Section \ref{sec:influence-spn-performance}. We report the mean balanced accuracy and standard deviation on the test set averaged across 25 runs. Figure \ref{fig:ablation_spn} is obtained by taking the difference between the right and middle columns of the numbers presented in this table.}

\centering
\begin{tabular}{@{}lrr@{}}
\toprule
Dataset          & \textbf{WPFS without SPN} & \textbf{WPFS with SPN} \\ \midrule
cll              & $75,65 \pm 9,17$         & $76,05 \pm 7,26$      \\
lung             & $94,32 \pm 5,34$         & $94,65 \pm 4,83$      \\
metabric-dr      & $57,28 \pm 6,62$         & $59,05 \pm 8,62$      \\
metabric-pam50   & $93,67 \pm 6,43$         & $95,5 \pm 4,8$        \\
prostate         & $86,96 \pm 6,73$         & $89 \pm 7,62$         \\
smk              & $63,4 \pm 8,19$          & $66,08 \pm 6,09$      \\
tcga-2ysurvival  & $56,34 \pm 9,54$         & $56,52 \pm 6,24$      \\
tcga-tumor-grade & $54,44 \pm 10,48$        & $55,29 \pm 11,29$     \\
toxicity         & $83,99 \pm 7,01$         & $88,01 \pm 6,31$      \\ \bottomrule
\end{tabular}
\end{table}

\section{Ablation Weight Predictor Network on the Classification Accuracy}
\label{appendix:ablation-wpn}

\begin{figure*}[h!]
    \centering
    \includegraphics[width=0.7\linewidth]{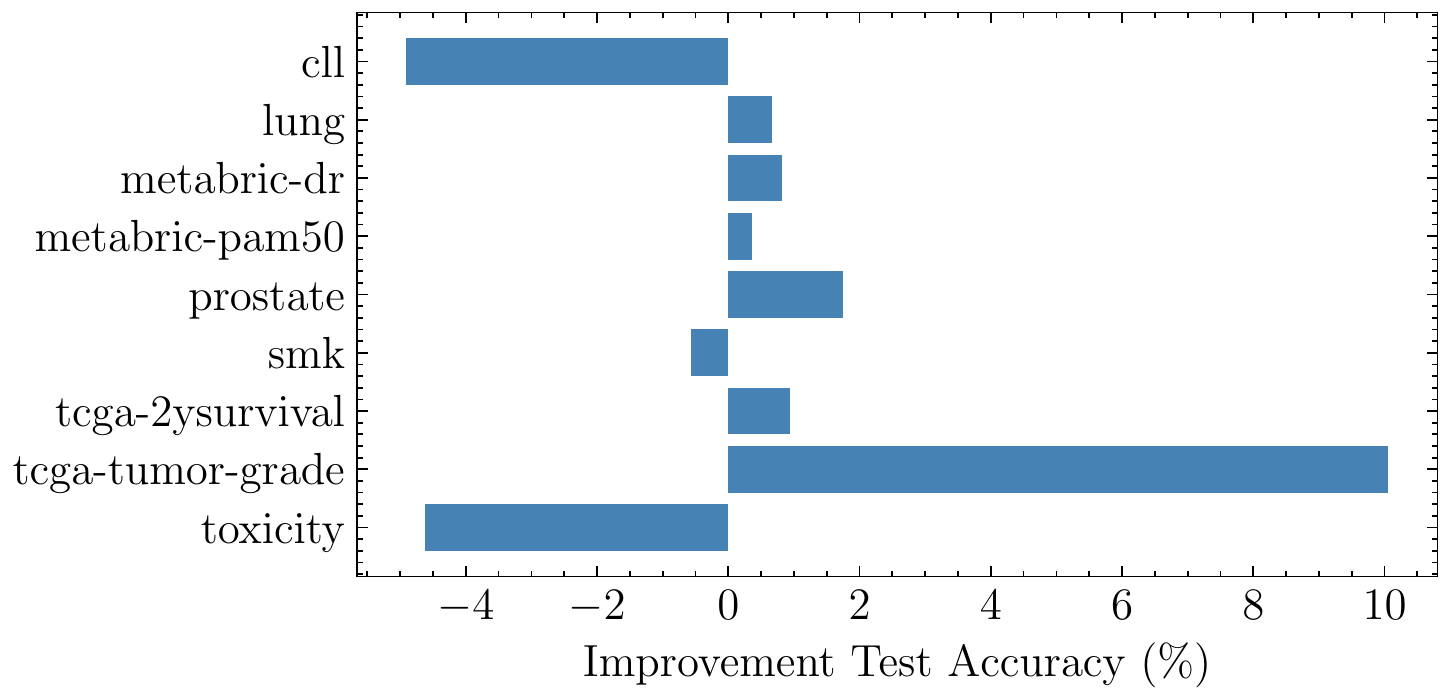}
    \caption{Ablation on including the WPN in WPFS. We trained a WPFS model for each dataset with and without the WPN. We tuned the sparsity hyper-parameter $\lambda \in \{0, 3e-5\}$ for each dataset and selected the best-performing model based on the validation cross-entropy. The x-axis represents the absolute improvement in the test accuracy by including the WPN. The WPN improves the overall performance across most datasets, except in the cases of ‘cll’ and ‘toxicity’ when it exhibits subpar but comparable performance.}
    \label{fig:ablation_wpn}
\end{figure*}

\clearpage
\section{Ablation Sparsity Loss Hyper-parameter $\lambda$ on the Classification Accuracy}
\label{appendix:varying-lambda-performance}

\begin{table*}[h!]
\centering
\caption{Test accuracy of WPFS trained with different sparsity loss hyper-parameter $\lambda$. We report the mean balanced accuracy and standard deviation, averaged over 25 runs.}
\label{tab:varying-lambda-performance}

\begin{tabular}{@{}lrrrrrr@{}}
\toprule
                 Sparsity parameter & $\lambda = 0$                 & $\lambda = 3e-6$         & $\lambda = 3e-5$           & $\lambda = 3e-4$            & $\lambda = 3e-3$ &  $\lambda = 1e-2$                    \\ \midrule
cll              & $76,05 \pm 7,26$  & $78,06 \pm 7,98$ & $78,27 \pm 10,05$ & $79,14 \pm 4,45$  & $77,5 \pm 7,35$ & $75,49 \pm 9,77$ \\
lung             & $94,65 \pm 4,83$  & $94,09 \pm 4,81$ & $94,83 \pm 4,2$   & $93,36 \pm 5,76$  & $93,89 \pm 4,99$ & $93,56 \pm 4,96$ \\
metabric-dr      & $59,05 \pm 8,62$  & $58,9 \pm 6,99$  & $56,89 \pm 8,4$   & $57,63 \pm 10,61$ & $54,6 \pm 8,48$ & $53,13 \pm 7,4$ \\
metabric-pam50  & $95,5 \pm 4,8$   & $95,96 \pm 4,11$ & $95,86 \pm 4,55$ & $94,63 \pm 4,64$ & $92,76 \pm 7,78$ & $92,45 \pm 6,03$ \\
prostate         & $89 \pm 7,62$ & $89.11 \pm 6.53$    & $87,84 \pm 6,26$ & $88,42 \pm 5,69$  & $89,15 \pm 6,73$  & $ 86.77 \pm 5.39 $      \\
smk              & $66,08 \pm 6,09$  & $63,91 \pm 7,43$ & $66,89 \pm 6,21$  & $64,16 \pm 6,45$  & $65,84 \pm 6,09$ & $62,73 \pm 8,63$\\
tcga-2ysurvival & $56,52 \pm 6,24$ & $57,04 \pm 6,66$ & $59,54 \pm 6,93$ & $57,03 \pm 5,89$ & $57,44 \pm 5,81$ & $56,41 \pm 7.55$ \\
tcga-tumor-grade & $55,29 \pm 11,29$ & $51,99 \pm 9.04$ & $55,91 \pm 8,57$ & $52,25 \pm 10,19$ & $51,78 \pm 9,86$  & $53,23 \pm 11.47$  \\
toxicity         & $88,01 \pm 6,31$  & $86,43 \pm 5,38$ & $88,29 \pm 5,27$  & $83,83 \pm 6,98$  & $86,8 \pm 7,02$ & $82,78 \pm 7,11$ \\ \bottomrule
\end{tabular}%
\end{table*}

\section{Ablation Varying Dataset Size}
\label{appendix:ablation-dataset-size}

We performed ablations with varying numbers of samples from the `metabric-dr' and `metabric-pam50' datasets. We trained WPFS and MLP using the settings presented in Appx. \ref{appendix:hyperparam}. For WPFS, we used the best sparsity hyper-parameter for dataset size $200$ ($\lambda=0$ for `metabric-dr' and $\lambda=3e-6$ for `metabric-pam50'), and we did not tune it for different sizes of the dataset -- though that is likely to improve the performance of WPFS. For each dataset, we perform $5$-fold cross-validation repeated $5$ times, resulting in $25$ runs. We randomly select $10\%$ of the training data for validation for each fold, and we report the mean and standard deviation of the balanced accuracy across all $25$ runs.

\begin{figure*}[h!]
    \centering
    \includegraphics[width=0.9\linewidth]{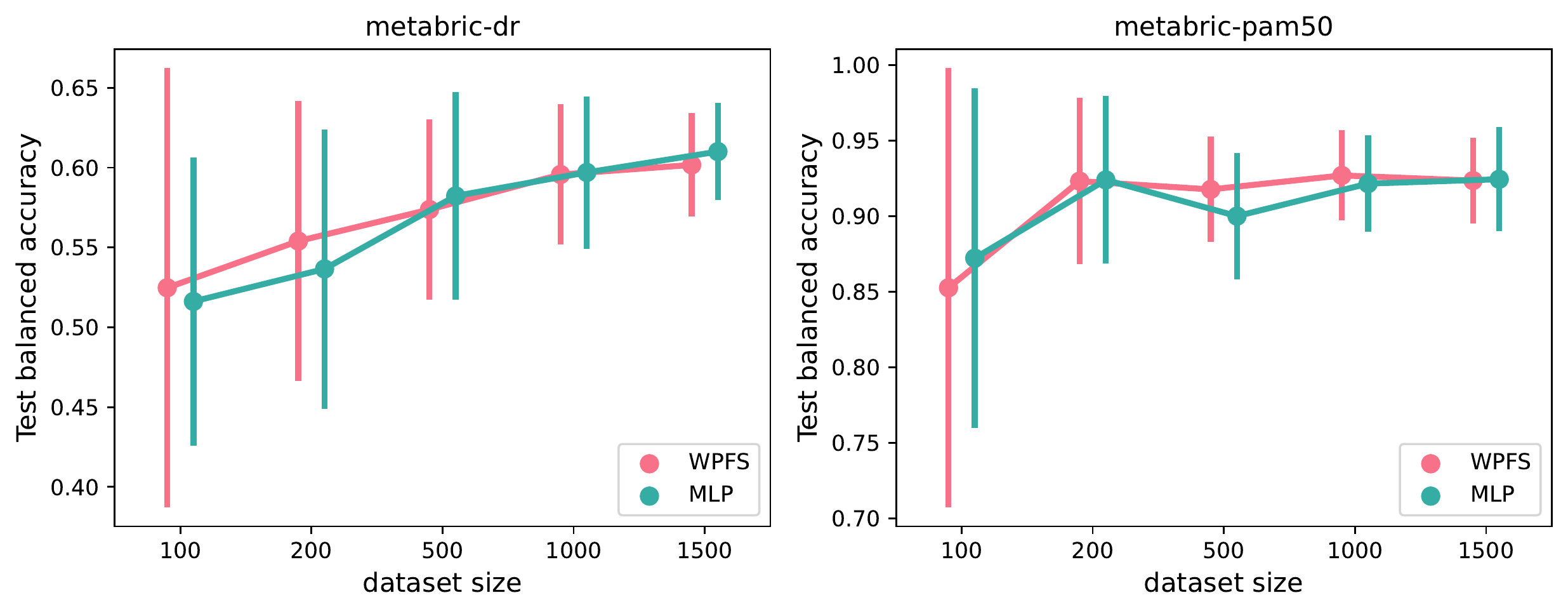}
    \caption{Test performance of WPFS and MLP on two datasets of varying sizes. We report the mean and standard deviation over 25 runs. Left: On `metabric-dr' -- a task on which all models obtain small accuracy -- we found that WPFS outperforms an MLP with very few samples. As the dataset increases, the instability decreases, and the performance of the two models becomes comparable. Right: The models perform comparably, likely because `metabric-pam50' is an easy task on which all models obtain high accuracy.}
    \label{fig:vary_dataset_size_test_acc}
\end{figure*}

Figure \ref{fig:vary_dataset_size_test_acc} compares the test accuracy of WPFS with an MLP. On ‘metabric-dr’, we found that WPFS outperforms an MLP with very few samples, but the performance of the two models converges as the dataset size increases. 

\begin{figure*}[h!]
    \centering
    \includegraphics[width=0.9\linewidth]{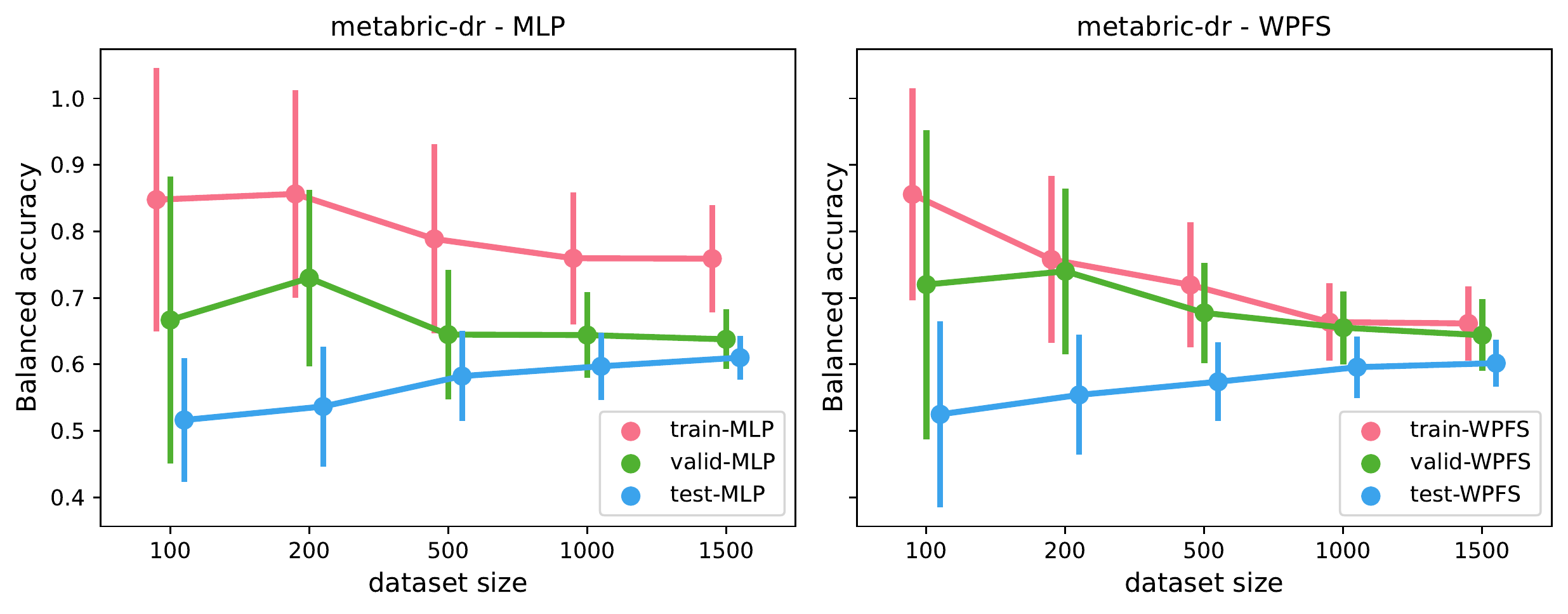}
    \caption{The train/valid/test performance of an MLP (left) and WPFS (right) on `metabric-dr'. We report the mean and standard deviation accuracy over 25 runs. As the dataset size increases, the generalisation gap between train-valid decreases, as it becomes harder to overfit a large training dataset. The test performance increases with the dataset size, and the instability decreases. Although the test performance of the two models converges for large dataset sizes, the generalisation gap of WPFS is smaller, indicating smaller overfitting.}
    \label{fig:vary_dataset_size_metabric_dr}
\end{figure*}

\begin{figure*}[h!]
    \centering
    \includegraphics[width=0.9\linewidth]{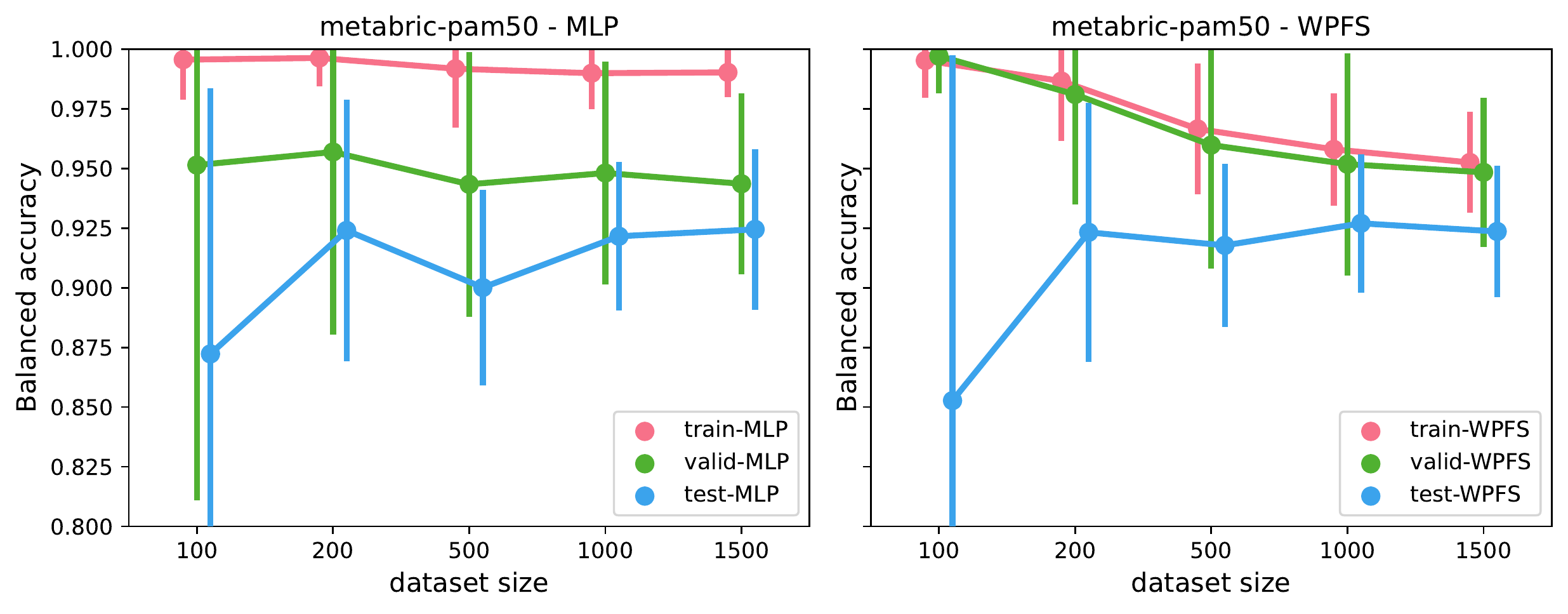}
    \caption{The train/valid/test performance of an MLP (left) and WPFS (right) on `metabric-pam50'. We report the mean and standard deviation accuracy over 25 runs. As the dataset size increases, the instability decreases, and the generalisation gap between train-valid decreases as it becomes harder to overfit a large training dataset. The test performance remains constant from around 200 samples, likely because the models already reach high test accuracy. Although the test performance of the two models converges for large dataset sizes, the generalisation gap of WPFS is smaller, indicating smaller overfitting.}
    \label{fig:vary_dataset_size_metabric_pam50}
\end{figure*}

For completeness, we include the train/valid/test accuracy of WPFS and MLP on both datasets: `metabric-dr' in Figure \ref{fig:vary_dataset_size_metabric_dr} and `metabric-pam50' in Figure \ref{fig:vary_dataset_size_metabric_pam50}. As the dataset size increases, the generalisation gap between train-valid decreases, as it becomes harder to overfit a large training dataset. The test performance increases with the dataset size, and the instability decreases. Although the test performance of the two models converges for large dataset sizes, the generalisation gap of WPFS is smaller, indicating smaller overfitting.

\clearpage
\section{Complete Results on Classification Accuracy}
\label{sec:full-results-additional-baselines}

\begin{table}[h!]
\centering
\caption{Evaluation on $9$ real-world biomedical datasets. We report the mean balanced accuracy and standard deviation averaged over $25$ runs.}
\label{tab:full-results}

\resizebox{\textwidth}{!}{%
\begin{tabular}{@{}llllllllll@{}}
\toprule
Method &
  cll &
  lung &
  metabric-dr &
  metabric-pam50 &
  prostate &
  smk &
  tcga-2ysurvival &
  tcga-tumor-grade &
  toxicity \\ \midrule

DietNetworks &
  $68.84 \pm 9.21$ &
  $90.43 \pm 6.23$ &
  $56.98 \pm 8.70$ &
  $95.02 \pm 4.77$ &
  $81.71 \pm 11.04$ &
  $62.71 \pm 9.38$ &
  $53.62 \pm 5.46$ &
  $46.69 \pm 7.11$ &
  $82.13 \pm 7.42$ \\
FsNet &
  $  66.38 \pm 9.22 $ &
  $  91.75 \pm 3.04 $ &
  $56.92 \pm 10.13$ &
  $  83.86 \pm 8.16 $ &
  $  84.74 \pm 9.80 $ &
  $  56.27 \pm 9.23 $ &
  $  53.83 \pm 7.94 $ &
  $  45.94 \pm 9.80 $ &
  $  60.26 \pm 8.10 $ \\
CAE &
  $  71.94 \pm 13.45 $ &
  $  85 \pm 5.04 $ &
  $57.35 \pm 9.37$ &
  $  95.78 \pm 3.61 $ &
  $  87.6 \pm 7.84 $ &
  $  59.96 \pm 10.98 $ &
  $  52.79 \pm 8.34 $ &
  $  40.69 \pm 7.36 $ &
  $  60.36 \pm 11.29 $ \\
LassoNet &
  $30.63 \pm 8.68$ &
  $25.11 \pm 9.81$ &
  $48.88 \pm 5.74$ &
  $48.81 \pm 10.82$ &
  $54.78 \pm 10.58$ &
  $51.04 \pm 8.56$ &
  $46.08 \pm 9.21$ &
  $33.49 \pm 7.55$ &
  $26.67 \pm 8.66$ \\
DNP &
  $85.12 \pm 5.46$ &
  $92.83 \pm 5.65$ &
  $55.79 \pm 7.06$ &
  $93.56 \pm 5.54$ &
  $90.25 \pm 5.98$ &
  $66.89 \pm 7.64$ &
  $58.13 \pm 8.2$ &
  $44.71 \pm 5.98$ &
  $93.5 \pm 6.17$ \\
SPINN &
  $85.34 \pm 5.47$ &
  $92.26 \pm 6.68$ &
  $56.13 \pm 7.24$ &
  $93.56 \pm 5.54$ &
  $89.27 \pm 5.91$ &
  $68.43 \pm 7.99$ &
  $57.70 \pm 7.07$ &
  $44.28 \pm 6.84$ &
  $93.5 \pm 4.85$ \\
TabNet &
  $57.81 \pm 9.97 $ &
  $77.65 \pm 12.92$ &
  $49.18 \pm 9.60$ &
  $83.60 \pm 11.44 $ &
  $65.66 \pm 14.75$ &
  $54.57 \pm 8.72 $ &
  $ 51.58 \pm 9.90$ &
  $39.34 \pm 7.92$ &
  $40.06 \pm 11.37$ \\
MLP &
  $78.30 \pm 8.98$ &
  $94.20 \pm 4.94$ &
  $59.56 \pm 5.50$ &
  $94.31 \pm 5.39$ &
  $88.76 \pm 5.55$ &
  $64.42 \pm 8.44$ &
  $56.28 \pm 6.72$ &
  $48.19 \pm 7.75$ &
  $93.21 \pm 6.13$ \\
Random Forest &
  $82.06 \pm 6.51$ &
  $91.81 \pm 6.94$ &
  $51.38 \pm 3.78$ &
  $89.11 \pm 6.59$ &
  $90.78 \pm 7.17$ &
  $68.16 \pm 7.56$ &
  $61.30 \pm 6.00$ &
  $50.93 \pm 8.48$ &
  $80.75 \pm 6.75$ \\
LightGBM &
  $85.59 \pm 6.51$ &
  $93.42 \pm 5.91$ &
  $58.23 \pm 8.55$ &
  $94.97 \pm 5.19$ &
  $91.38 \pm 5.71$ &
  $65.70 \pm 7.46$ &
  $57.08 \pm 7.86$ &
  $49.11 \pm 10.32$ &
  $82.40 \pm 6.48$ \\
  \midrule
  \textbf{WPFS (ours)} &
 \shortstack[l]{$79.14 \pm 4.45$ \\ $[\lambda=3e-4]$} &
  \shortstack[l]{$94.83 \pm 4.2$ \\ $[\lambda=3e-5]$ } &
  \shortstack[l]{$59.05 \pm 8.62 $ \\ $[\lambda=0]$} &
  \shortstack[l]{$95.96 \pm 4.11 $ \\ $[\lambda=3e-6]$} &
  \shortstack[l]{$89.15 \pm 6.73$ \\ $[\lambda=3e-3]$} &
  \shortstack[l]{$66.89 \pm 6.21$ \\ $[\lambda = 3e-5]$} &
  \shortstack[l]{$59.54 \pm 6.93$ \\ $[\lambda = 3e-5]$} &
  \shortstack[l]{$55.91 \pm 8.57$ \\ $[\lambda=3e-5]$} &
  \shortstack[l]{$88.29 \pm 5.27$ \\ $[\lambda=3e-5]$} \\

  \bottomrule
\end{tabular}%
}

\label{tab:my-table}
\end{table}

\section{Ablation Sparsity Loss Hyper-parameter $\lambda$ on the Feature Importance Scores}
\label{appendix:feature_importance_distribution_varying_lambda}

\begin{figure}[]
    \centering
    \includegraphics[width=0.55\textwidth]{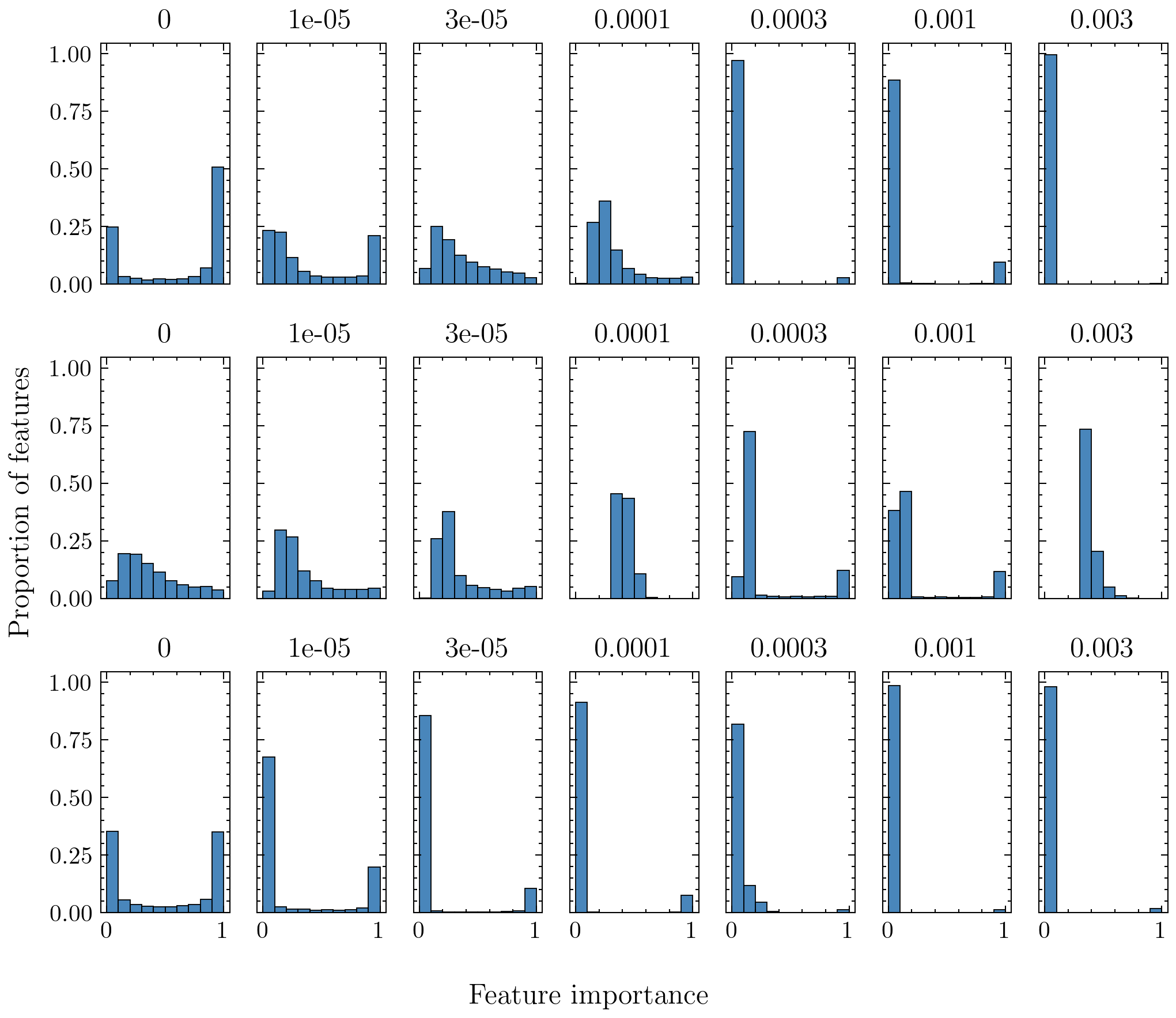}
    \caption{Feature importance scores for dataset \textbf{metabric-dr}, when increasing the sparsity hyper-parameters $\lambda$. Each row represents one randomly chosen training split, and each column represent a different sparsity hyper-parameter. Notice that increasing $\lambda$ leads to a larger proportion of features considered irrelevant (i.e., their feature importance scores tend to zero).}
    \label{fig:feature-importance-metabric-dr}
\end{figure}

\begin{figure}[]
    \centering
    \includegraphics[width=0.55\textwidth]{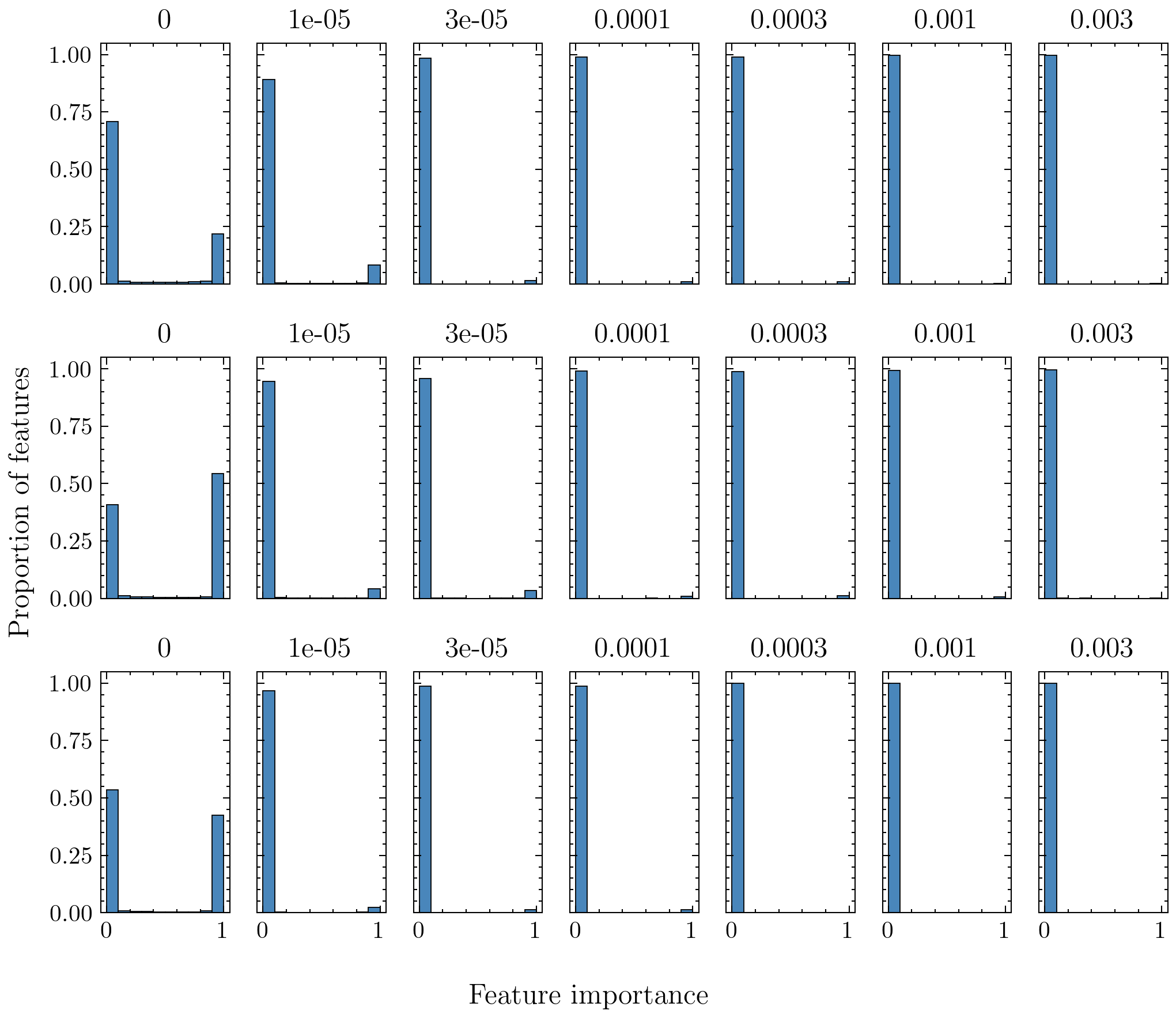}
    \caption{Feature importance for dataset \textbf{metabric-pam50}, when increasing the sparsity hyper-parameters $\lambda$. Each row represents one randomly chosen training split, and each column represent a different sparsity hyper-parameter. Notice that increasing $\lambda$ leads to a larger proportion of features considered irrelevant (i.e., their feature importance scores tend to zero).}
    \label{fig:feature-importance-pam50}
\end{figure}

We investigate the effect of the sparsity loss hyper-parameter $\lambda$ on the distribution of feature importance scores. For each dataset, we show three runs trained on different data splits. Across datasets, increasing $\lambda$ leads to fewer selected features. Notice that WPFS is uncertain which features are important in tasks with low test performance (e.g., Figure \ref{fig:feature-importance-metabric-dr}), it provides almost binary feature importance on tasks with high test performance (e.g., Figure \ref{fig:feature-importance-pam50}).

\begin{figure}[h!]
    \centering
    \includegraphics[width=0.53\textwidth]{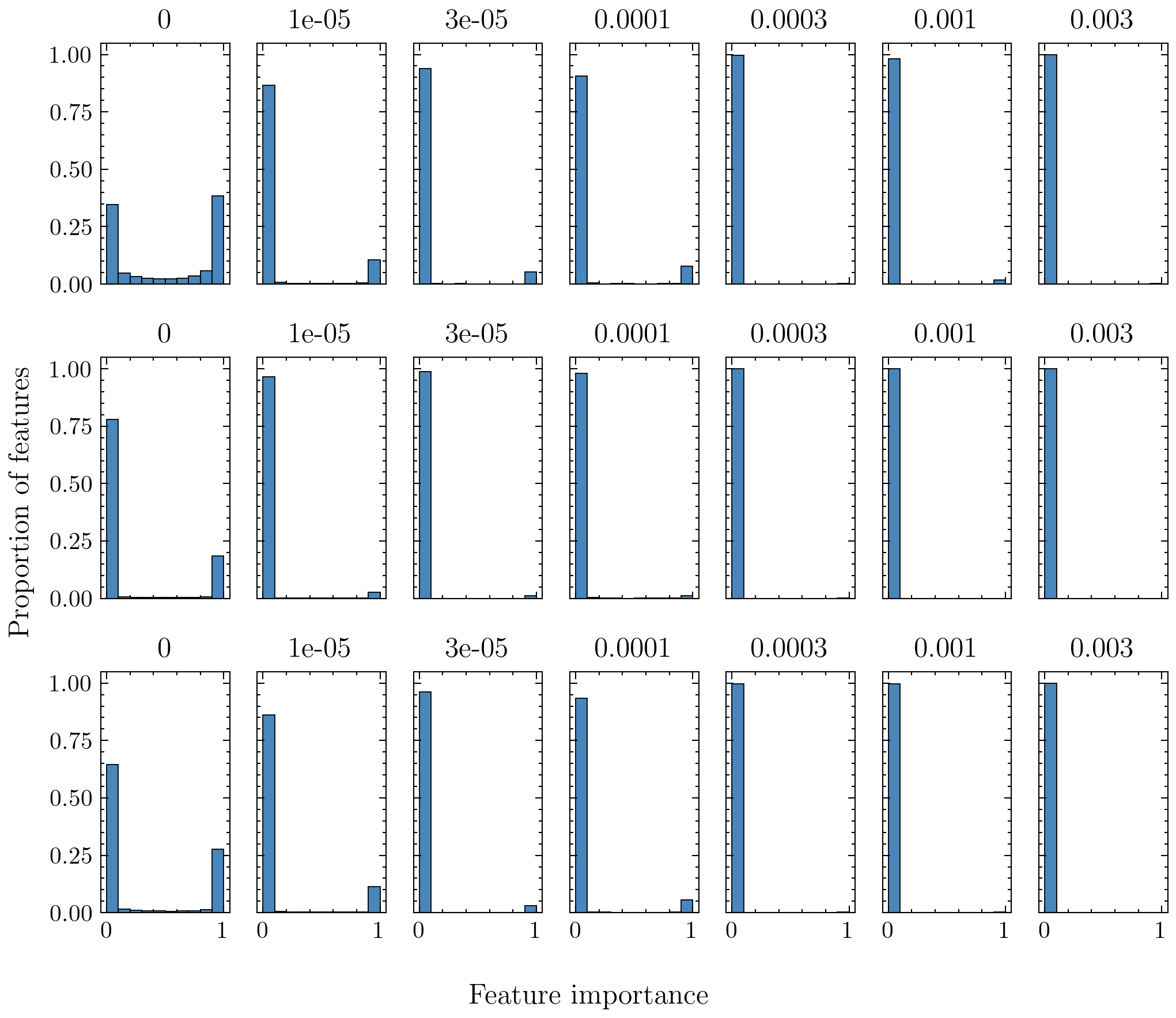}
    \caption{Feature importance for dataset \textbf{cll}, when increasing the sparsity hyper-parameters $\lambda$. Each row represents one randomly chosen training split, and each column represent a different sparsity hyper-parameter. Notice that increasing $\lambda$ leads to a larger proportion of features considered irrelevant (i.e., their feature importance scores tend to zero).}
\end{figure}

\begin{figure}[h!]
    \centering
    \includegraphics[width=0.53\textwidth]{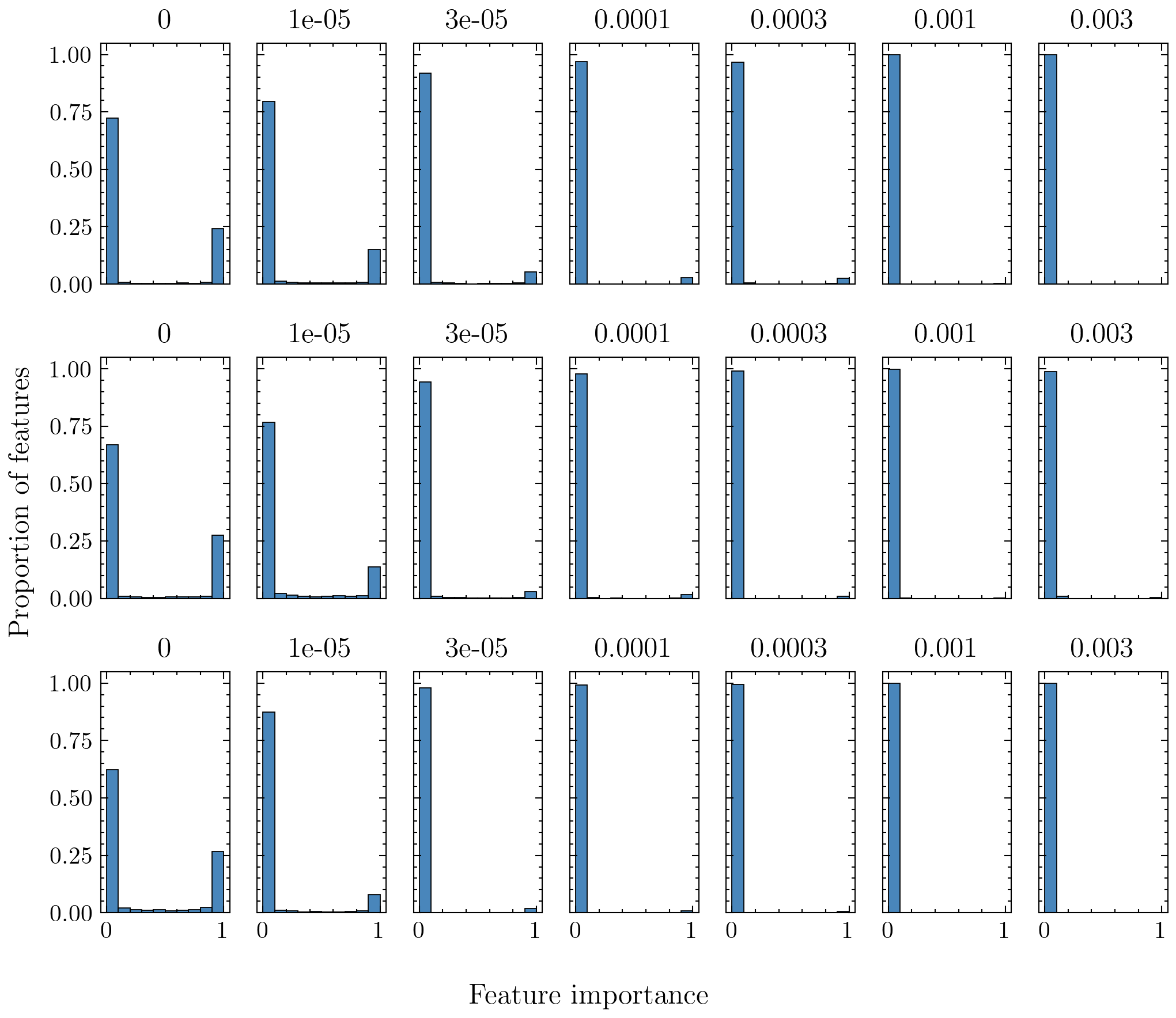}
    \caption{Feature importance scores for dataset \textbf{lung}, when increasing the sparsity hyper-parameters $\lambda$. Each row represents one randomly chosen training split, and each column represent a different sparsity hyper-parameter. Notice that increasing $\lambda$ leads to a larger proportion of features considered irrelevant (i.e., their feature importance scores tend to zero).}
\end{figure}

\begin{figure}[]
    \centering
    \includegraphics[width=0.55\textwidth]{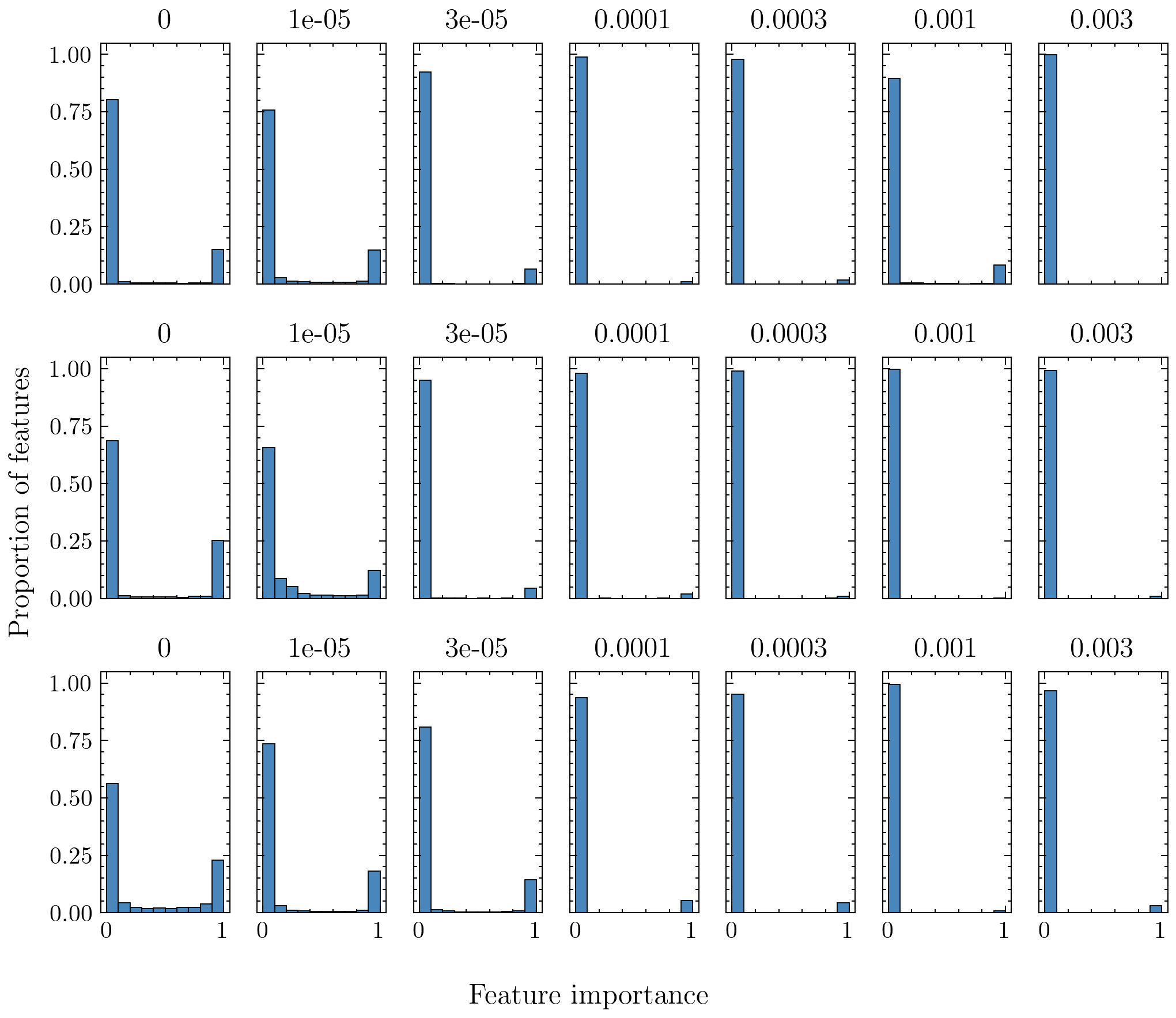}
    \caption{Feature importance scores for dataset \textbf{prostate}, when increasing the sparsity hyper-parameters $\lambda$. Each row represents one randomly chosen training split, and each column represent a different sparsity hyper-parameter. Notice that increasing $\lambda$ leads to a larger proportion of features considered irrelevant (i.e., their feature importance scores tend to zero).}
\end{figure}

\begin{figure}[]
    \centering
    \includegraphics[width=0.55\textwidth]{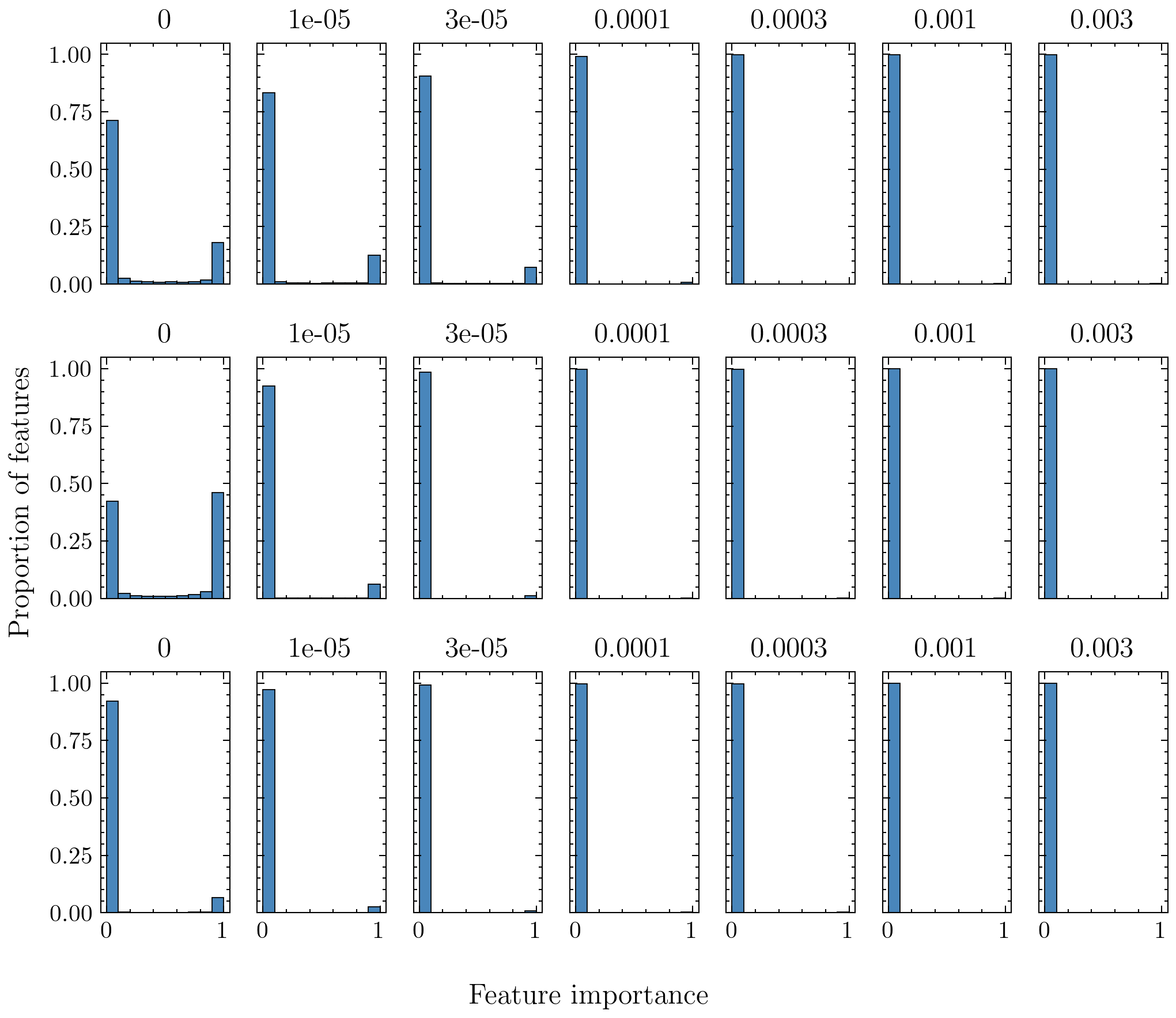}
    \caption{Feature importance scores for dataset \textbf{smk}, when increasing the sparsity hyper-parameters $\lambda$. Each row represents one randomly chosen training split, and each column represent a different sparsity hyper-parameter. Notice that increasing $\lambda$ leads to a larger proportion of features considered irrelevant (i.e., their feature importance scores tend to zero).}
\end{figure}

\begin{figure}[]
    \centering
    \includegraphics[width=0.55\textwidth]{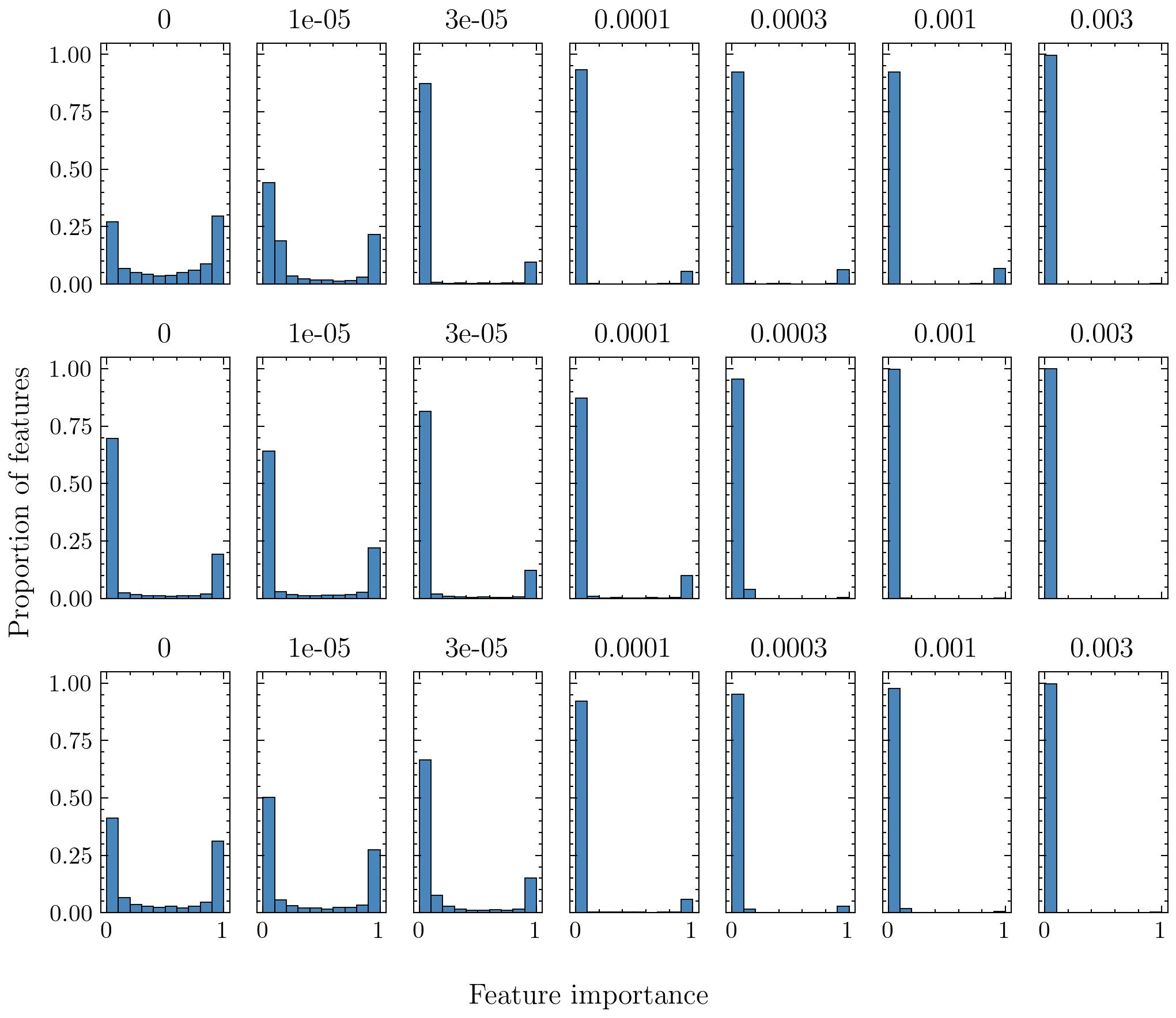}
    \caption{Feature importance scores for dataset \textbf{tcga-2ysurvival}, when increasing the sparsity hyper-parameters $\lambda$. Each row represents one randomly chosen training split, and each column represent a different sparsity hyper-parameter. Notice that increasing $\lambda$ leads to a larger proportion of features considered irrelevant (i.e., their feature importance scores tend to zero).}
\end{figure}

\begin{figure}[]
    \centering
    \includegraphics[width=0.55\textwidth]{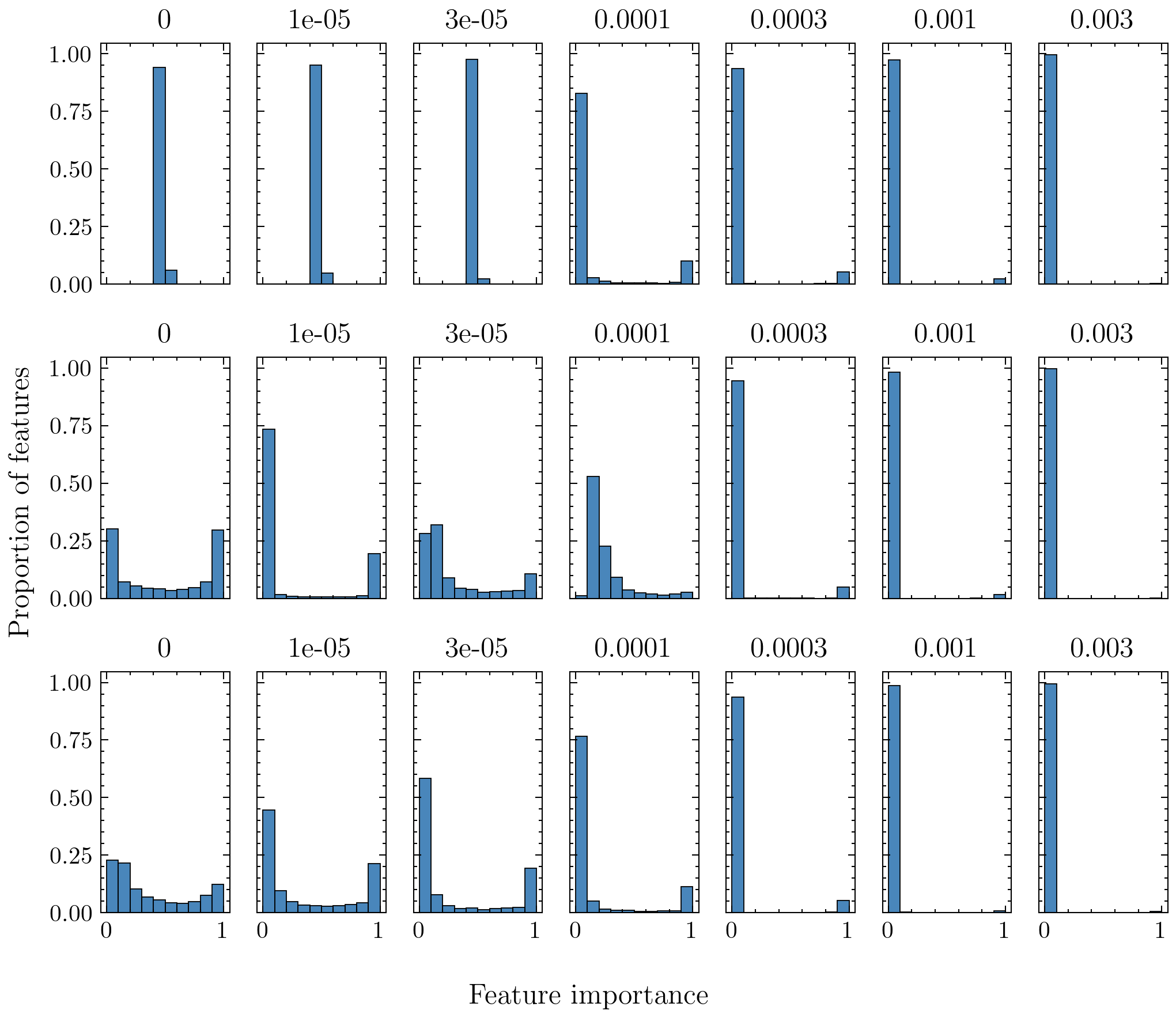}
    \caption{Feature importance scores for dataset \textbf{tcga-tumor-grade}, when increasing the sparsity hyper-parameters $\lambda$. Each row represents one randomly chosen training split, and each column represent a different sparsity hyper-parameter. Notice that increasing $\lambda$ leads to a larger proportion of features considered irrelevant (i.e., their feature importance scores tend to zero).}
\end{figure}

\begin{figure}[t!]
    \centering
    \includegraphics[width=0.55\textwidth]{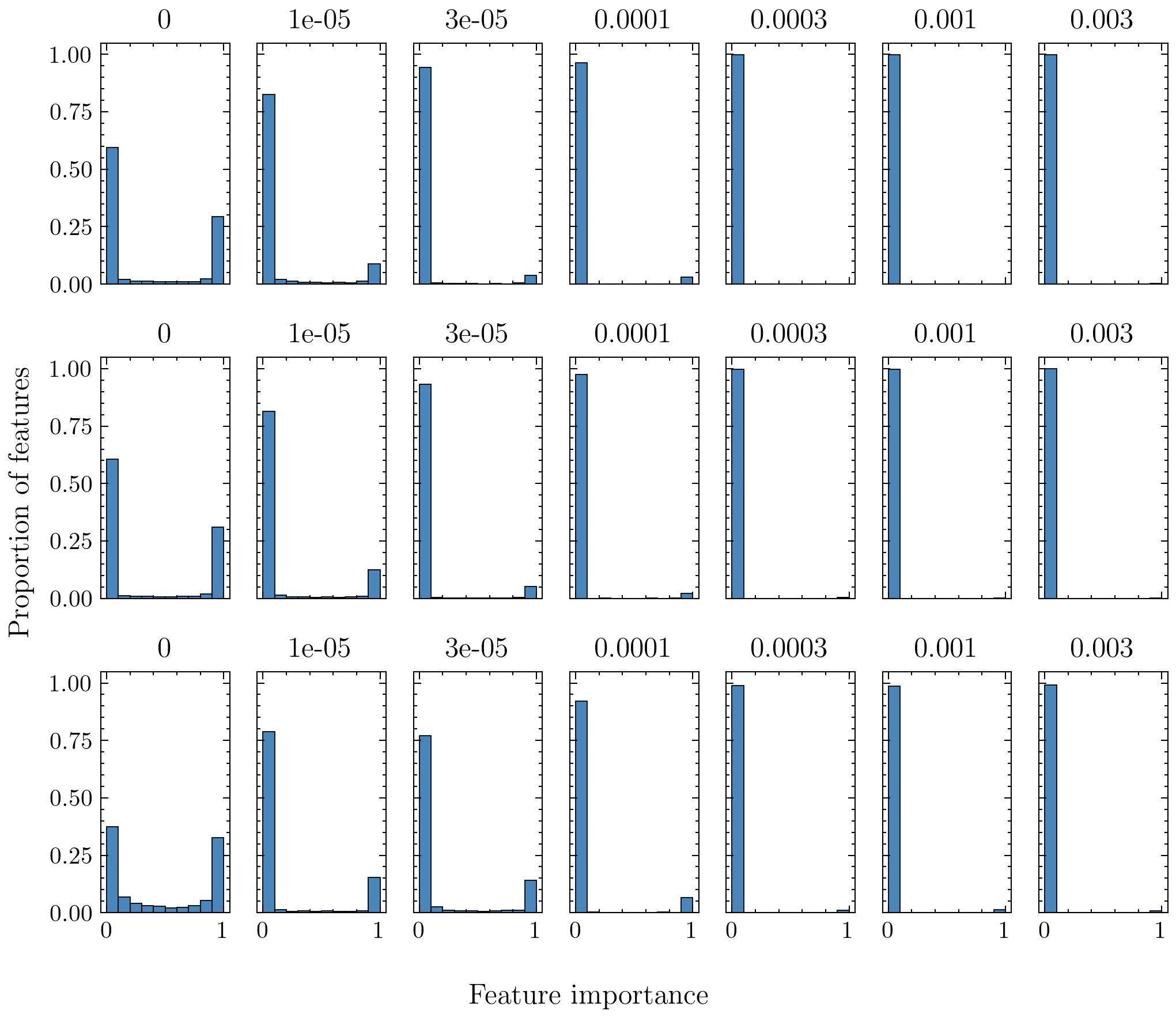}
    \caption{Feature importance scores for dataset \textbf{toxicity}, when increasing the sparsity hyper-parameters $\lambda$. Each row represents one randomly chosen training split, and each column represent a different sparsity hyper-parameter. Notice that increasing $\lambda$ leads to a larger proportion of features considered irrelevant (i.e., their feature importance scores tend to zero).}
\end{figure}

\clearpage
\section{Training Dynamics for WPFS and MLP}
\label{appendix:training_dynamics}

\subsection{Loss curves}

\begin{figure}[h!]
    \centering
    \includegraphics[width=0.8\textwidth]{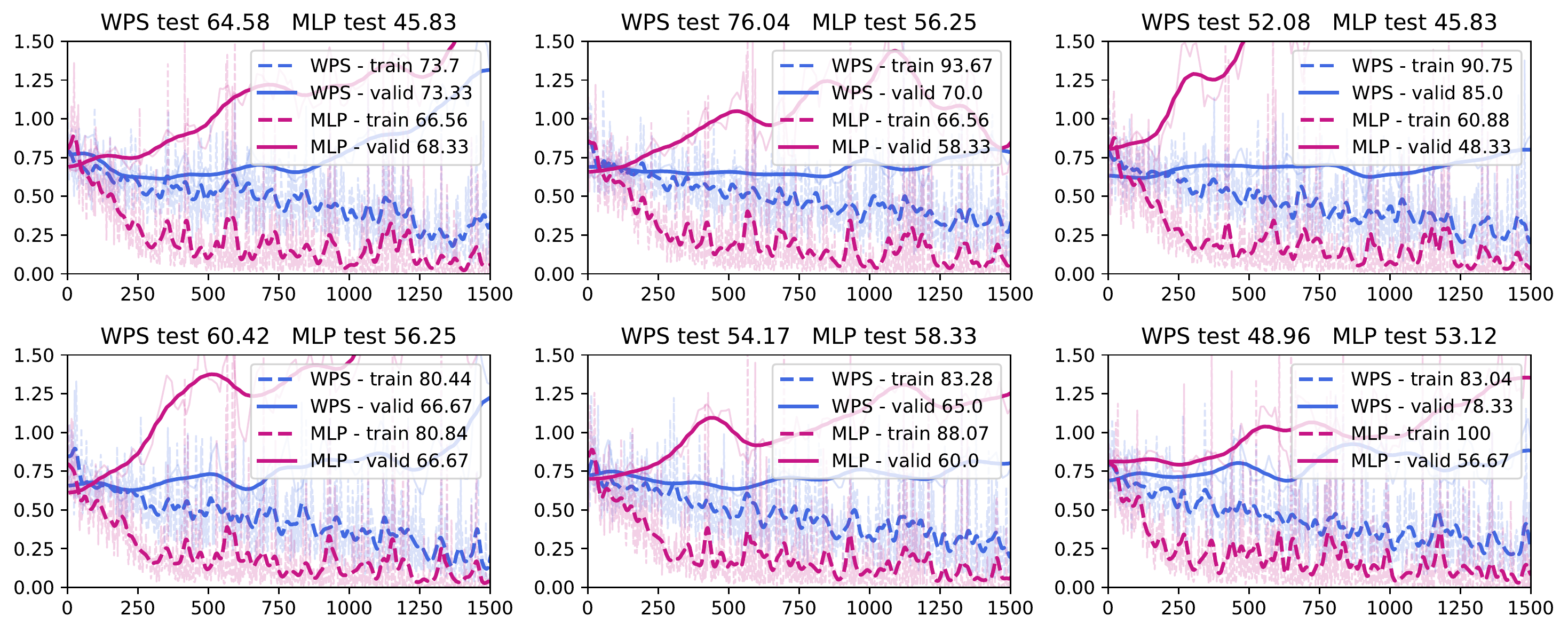}
    \caption{Loss curve trends for dataset \textbf{`tcga-2ysurvival'} for 9 randomly selected runs. The x-axis represents the iteration, and the y-axis represents the weighted cross-entropy. Each subfigure displayes the test accuracy in the title, and the train/validation accuracy in the legend.}
    \label{fig:loss-curve-tcga-2ysurvival}
\end{figure}

We show the loss curves trends of six randomly selected runs for each dataset. We selected the model with the smallest validation loss for each run, and we display its performance (test balanced accuracy in the title and the train/validation balanced accuracy in the legend). As a general trend, we observe that the loss curve is a good indicator of the validation performance. When one loss diverges, the corresponding model has lower validation accuracy (see Figure \ref{fig:loss-curve-tcga-2ysurvival}). However, there is not a strong correlation between validation and test performance. We believe this is expected because the datasets have few samples, and we use $10\%$ of data for validation and $20\%$ for testing.

\begin{figure}[!ht]
    \centering
    \includegraphics[width=0.8\textwidth]{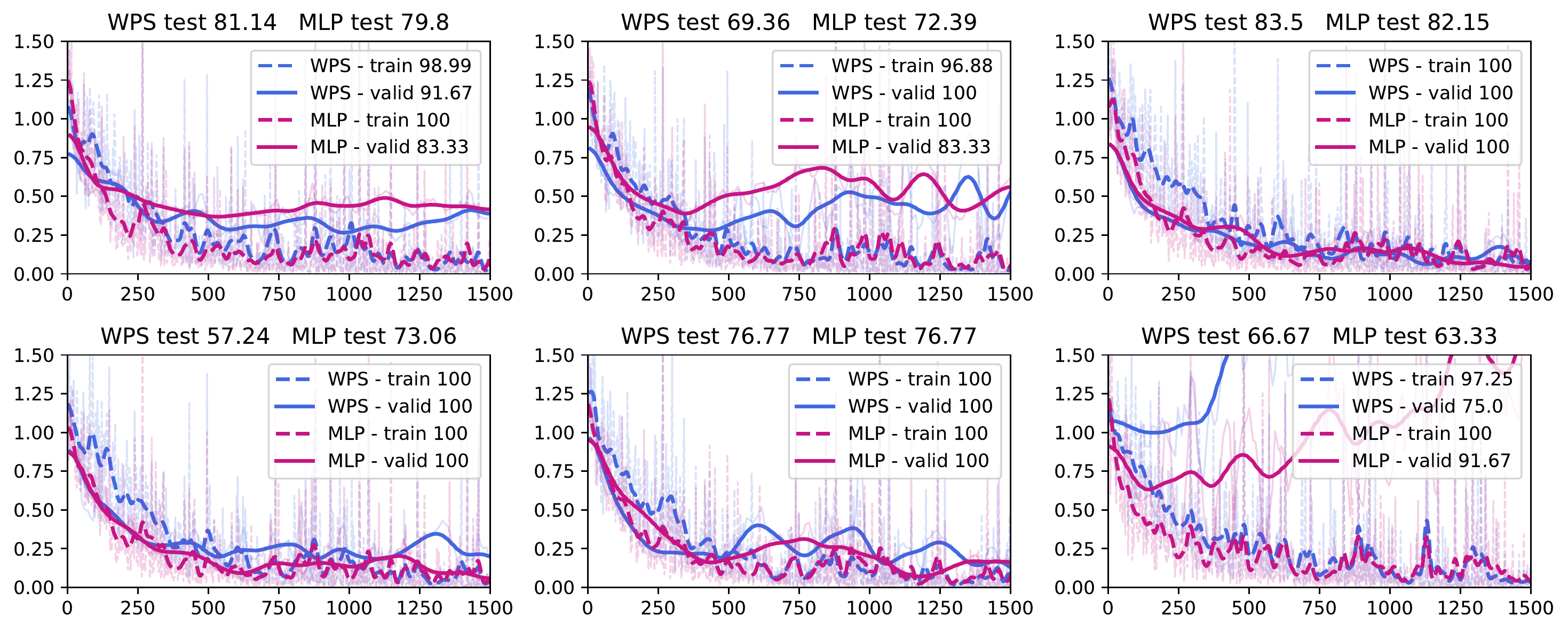}
    \caption{Loss curve trends for dataset \textbf{`cll'} for 9 randomly selected runs. The x-axis represents the iteration, and the y-axis represents the weighted cross-entropy. Each subfigure displayes the test accuracy in the title, and the train/validation accuracy in the legend.}
\end{figure}

\begin{figure}[!h]
    \centering
    \includegraphics[width=0.8\textwidth]{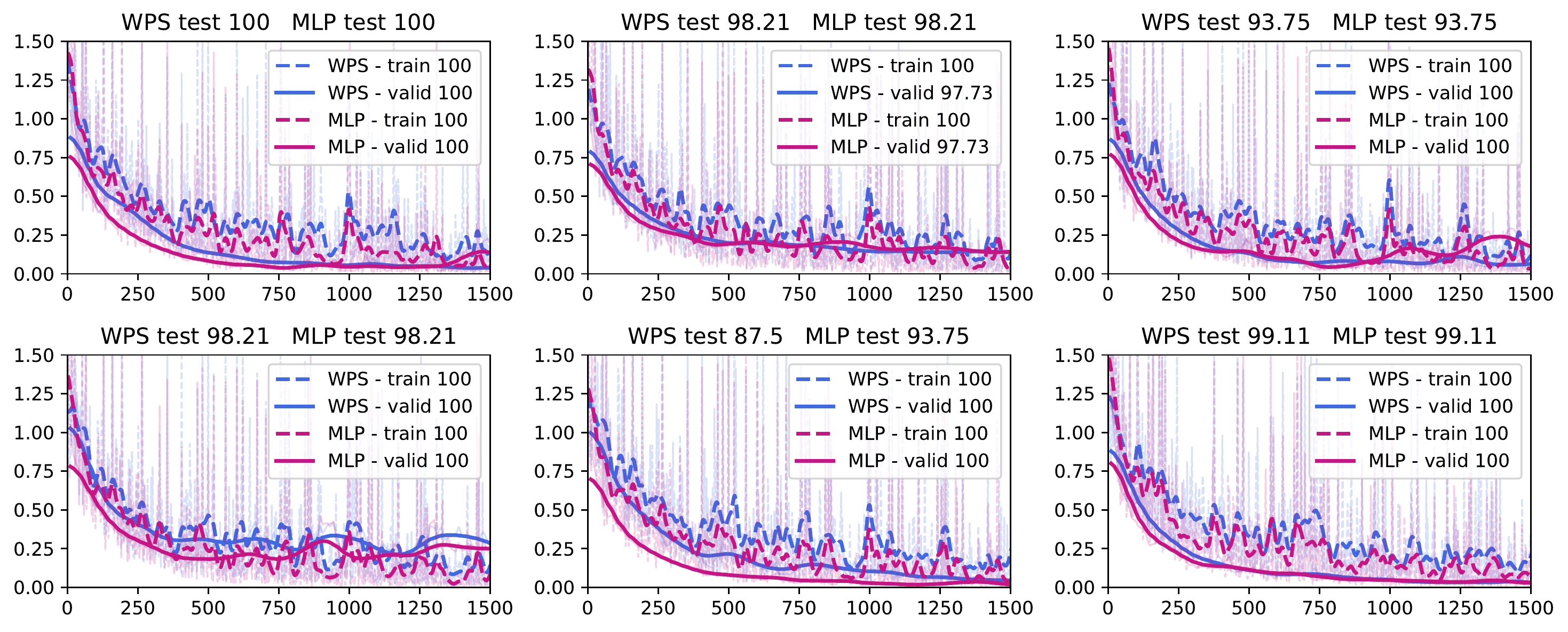}
    \caption{Loss curve trends for dataset \textbf{`lung'} for 9 randomly selected runs. The x-axis represents the iteration, and the y-axis represents the weighted cross-entropy. Each subfigure displayes the test accuracy in the title, and the train/validation accuracy in the legend.}
\end{figure}

\begin{figure}[!ht]
    \centering
    \includegraphics[width=0.8\textwidth]{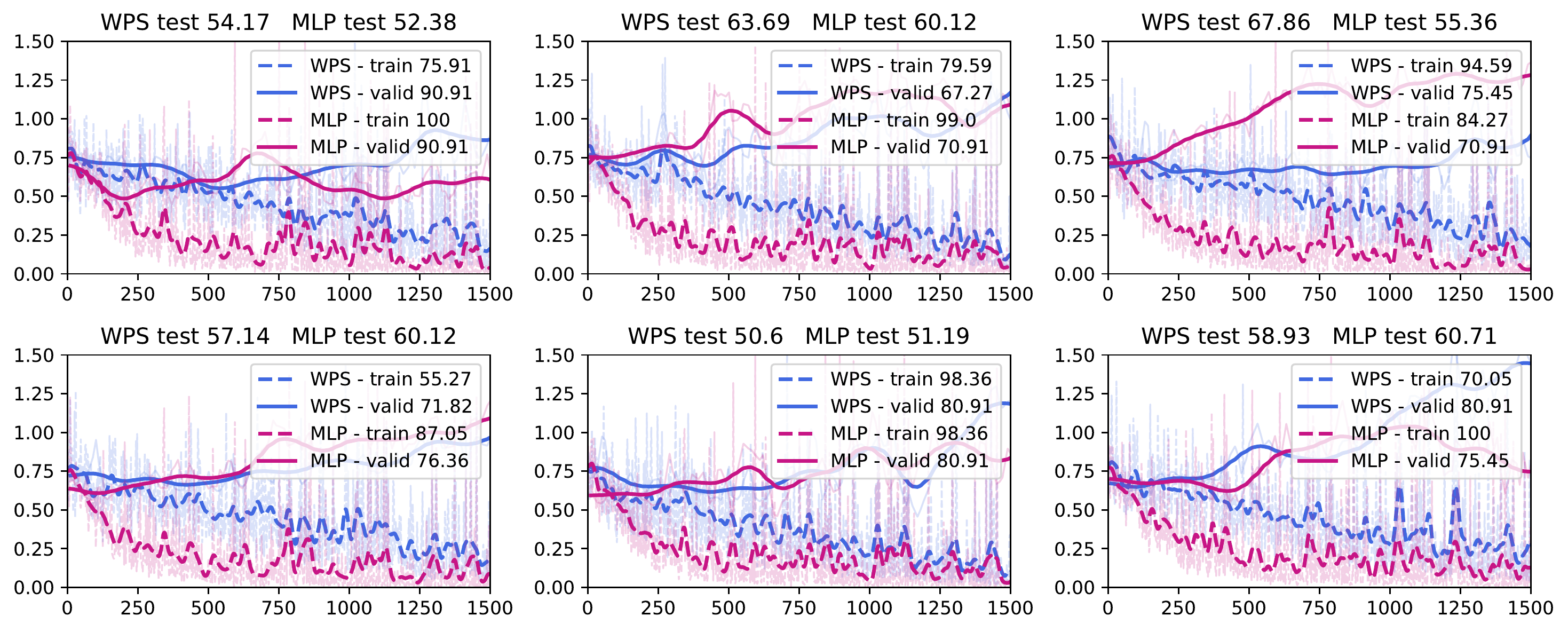}
    \caption{Loss curve trends for dataset \textbf{`metabric-dr'} for 9 randomly selected runs. The x-axis represents the iteration, and the y-axis represents the weighted cross-entropy. Each subfigure displayes the test accuracy in the title, and the train/validation accuracy in the legend.}
\end{figure}

\begin{figure}[h!]
    \centering
    \includegraphics[width=0.8\textwidth]{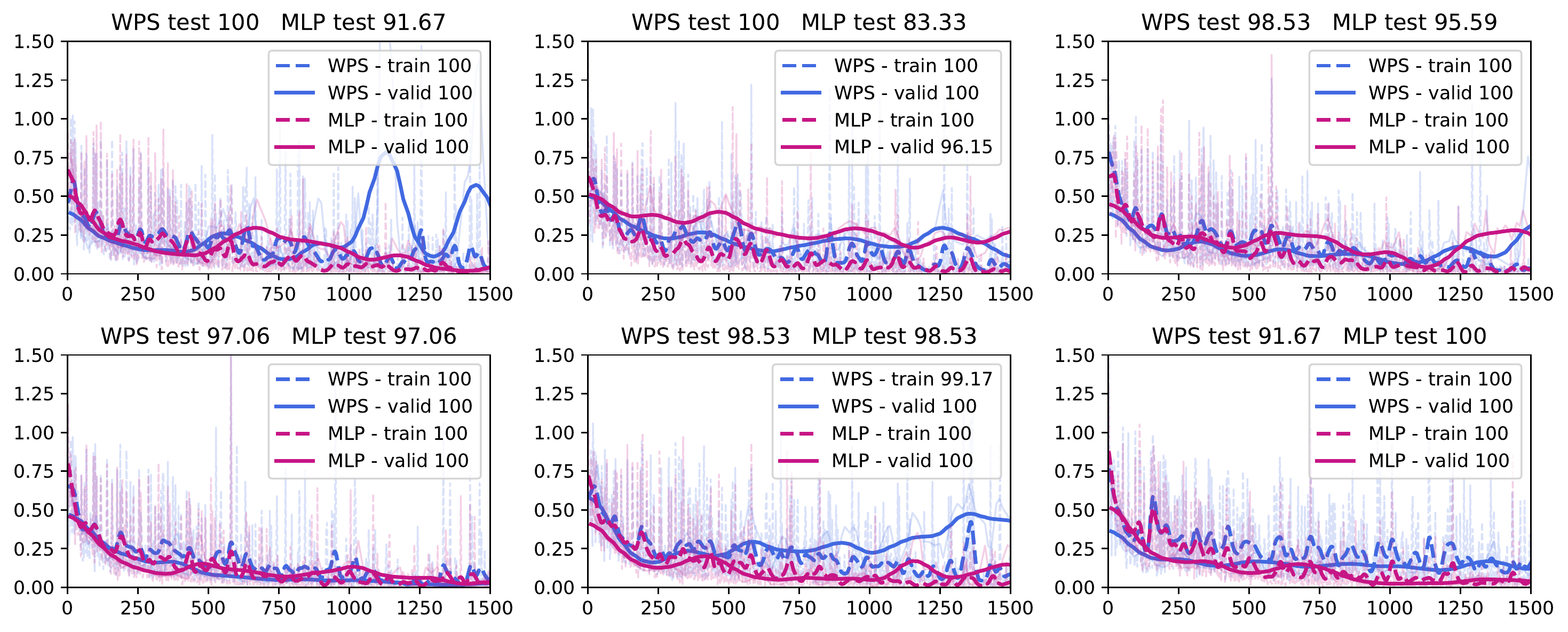}
    \caption{Loss curve trends for dataset \textbf{`metabric-pam50'} for 9 randomly selected runs. The x-axis represents the iteration, and the y-axis represents the weighted cross-entropy. Each subfigure displayes the test accuracy in the title, and the train/validation accuracy in the legend.}
\end{figure}

\begin{figure}[h!]
    \centering
    \includegraphics[width=0.8\textwidth]{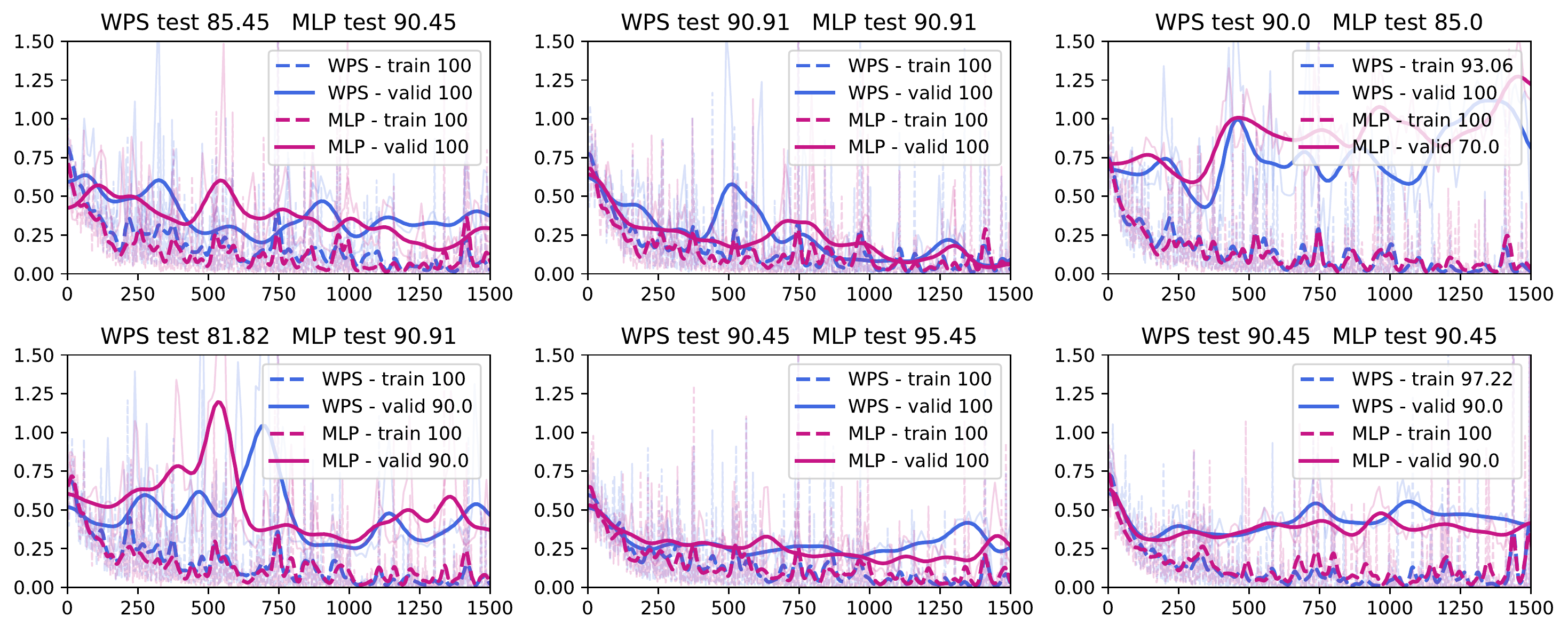}
    \caption{Loss curve trends for dataset \textbf{`prostate'} for 9 randomly selected runs. The x-axis represents the iteration, and the y-axis represents the weighted cross-entropy. Each subfigure displayes the test accuracy in the title, and the train/validation accuracy in the legend.}
\end{figure}

\begin{figure}[h!]
    \centering
    \includegraphics[width=0.8\textwidth]{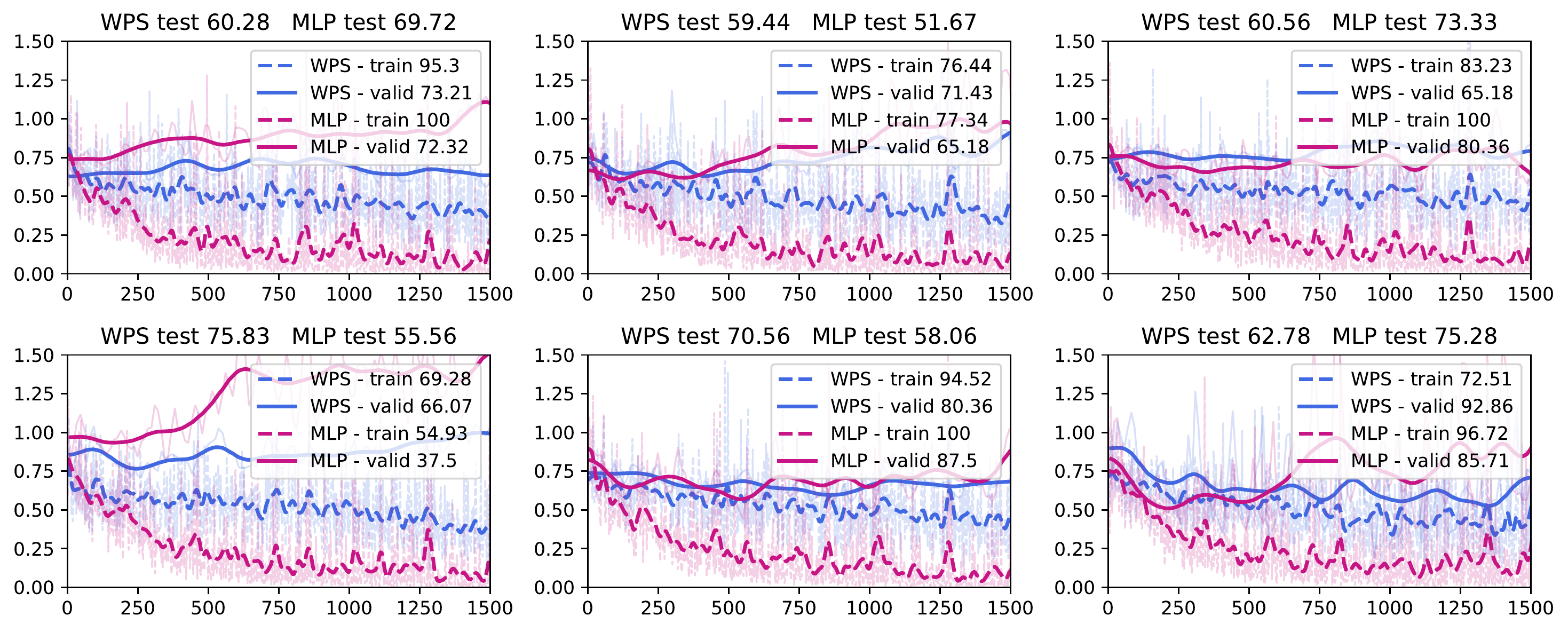}
    \caption{Loss curve trends for dataset \textbf{`smk'} for 9 randomly selected runs. The x-axis represents the iteration, and the y-axis represents the weighted cross-entropy. Each subfigure displayes the test accuracy in the title, and the train/validation accuracy in the legend.}
\end{figure}

\begin{figure}[h!]
    \centering
    \includegraphics[width=0.8\textwidth]{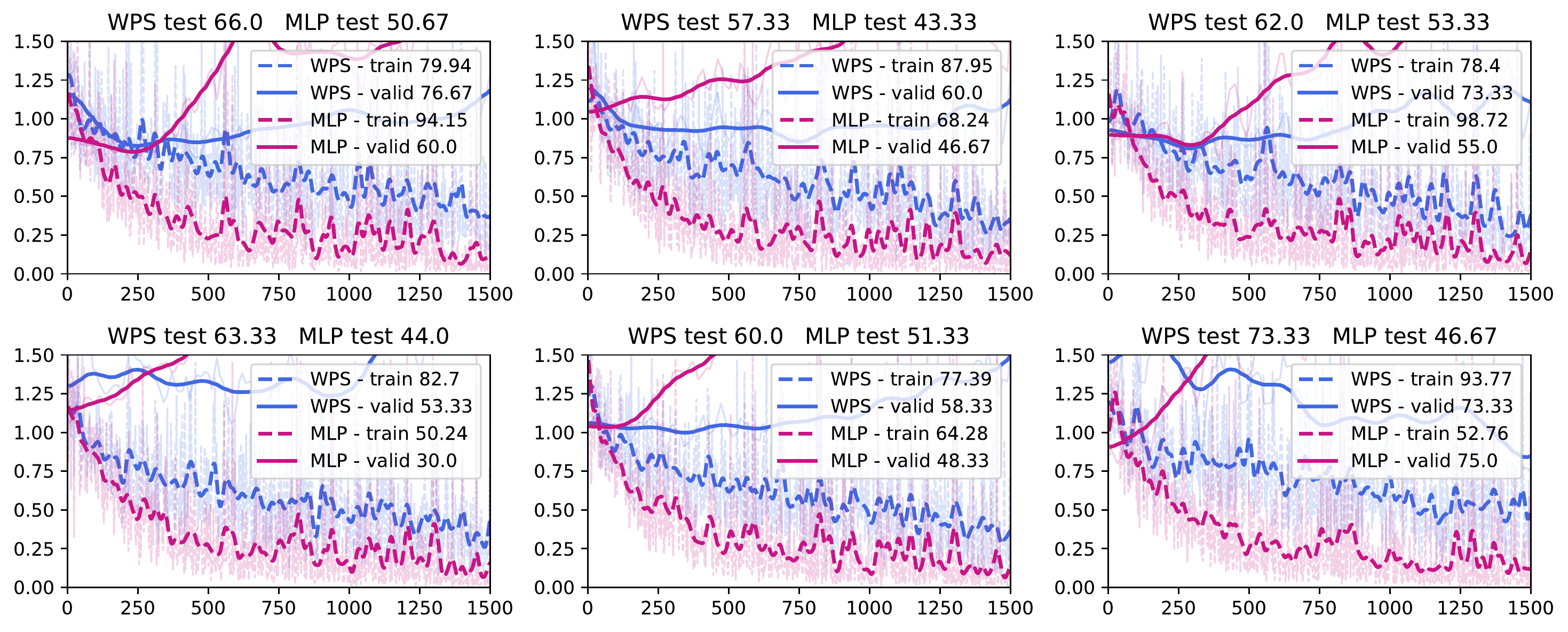}
    \caption{Loss curve trends for dataset \textbf{`tcga-tumor-grade'} for 9 randomly selected runs. The x-axis represents the iteration, and the y-axis represents the weighted cross-entropy. Each subfigure displayes the test accuracy in the title, and the train/validation accuracy in the legend.}
\end{figure}

\begin{figure}[t!]
    \centering
    \includegraphics[width=0.8\textwidth]{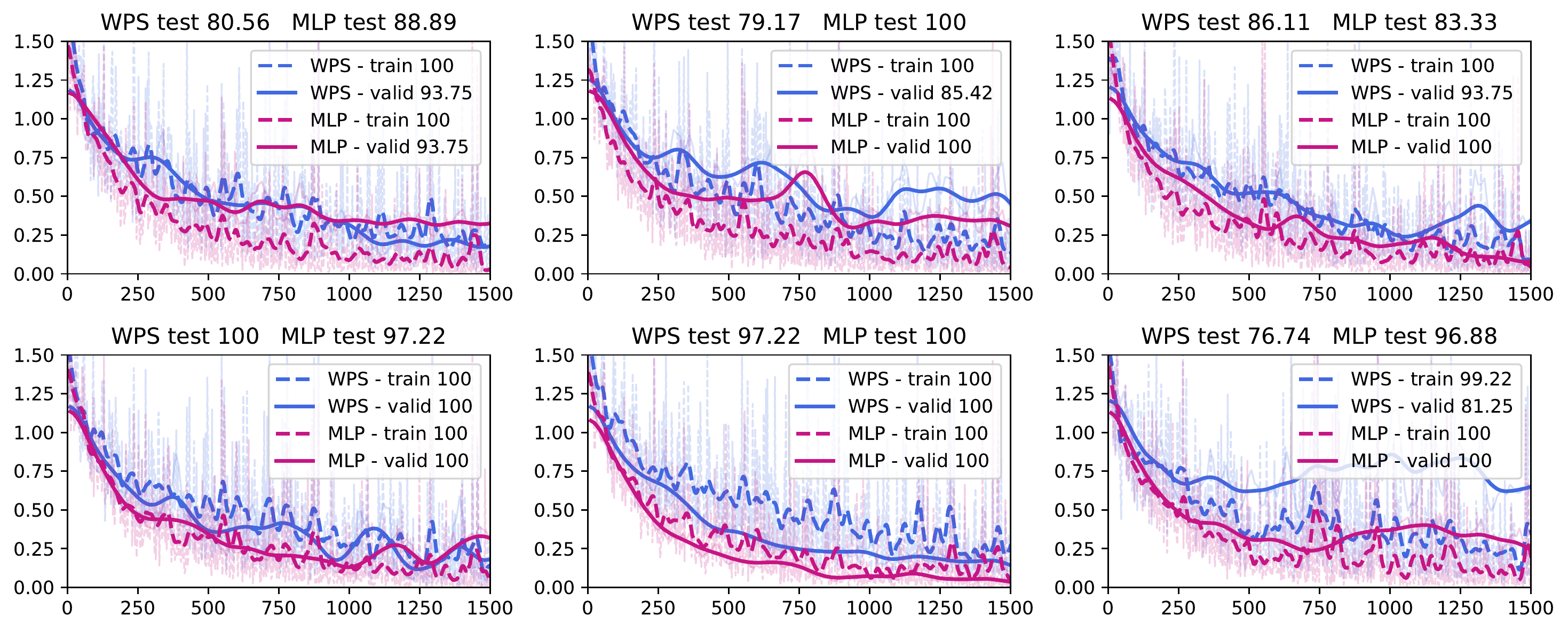}
    \caption{Loss curve trends for dataset \textbf{`toxicity'} for 9 randomly selected runs. The x-axis represents the iteration, and the y-axis represents the weighted cross-entropy. Each subfigure displayes the test accuracy in the title, and the train/validation accuracy in the legend.}
\end{figure}

\end{document}